\documentclass[sigconf]{acmart}
\usepackage{graphicx}
\usepackage{subcaption}
\usepackage{algpseudocode}
\usepackage{algorithm}
\usepackage{tcolorbox}
\usepackage{comment}
\usepackage{tcolorbox}
\AtBeginDocument{%
  }

\newcommand{\Foutse}[1]{\textcolor{red}{{\it [Foutse: #1]}}}

\newcommand{\Dima}[1]{\textcolor{blue}{{ [Dima says: #1] }}}

\copyrightyear{2024}
\acmYear{2024}
\setcopyright{acmlicensed}\acmConference[ASE '24]{39th IEEE/ACM International Conference on Automated Software Engineering }{October 27-November 1, 2024}{Sacramento, CA, USA}
\acmBooktitle{39th IEEE/ACM International Conference on Automated Software Engineering (ASE '24), October 27-November 1, 2024, Sacramento, CA, USA}
\acmDOI{10.1145/3691620.3695281}
\acmISBN{979-8-4007-1248-7/24/10}
\begin{document}

\title{In-Simulation Testing of Deep Learning Vision Models in Autonomous Robotic Manipulators
}

\author{Dmytro Humeniuk}
\orcid{0000-0002-2983-8312}
\affiliation{%
  \institution{Polytechnique Montréal}
 % \streetaddress{P.O. Box 1212}
  \city{Montreal}
  \state{Quebec}
  \country{Canada}
}
\email{dmytro.humeniuk@polymtl.ca}

\author{Houssem Ben Braiek}

\orcid{0000-0003-0542-9140}
\affiliation{%
  \institution{Sycodal}
 % \streetaddress{P.O. Box 1212}
  \city{Montreal}
  \state{Quebec}
  \country{Canada}
}
\email{h.benbraiek@sycodal.ca}

\author{Thomas Reid}
\orcid{0009-0001-3601-9323}
\affiliation{%
  \institution{Sycodal}
  \city{Montreal}
  \state{Quebec}
  \country{Canada}
}
%\email{t.reid@sycodal.ca}

\author{Foutse Khomh}
  \orcid{0000-0002-5704-4173}
\affiliation{%
  \institution{Polytechnique Montréal}
  \city{Montreal}
  \state{Quebec}
  \country{Canada}
}
%\email{foutse.khomh@polymtl.ca}
\renewcommand{\shortauthors}{Humeniuk et al.}
%%
%% Article type: Research, Review, Discussion, Invited or position
\acmArticleType{Research}
%%
%% Links to code and data
%\acmCodeLink{https://github.com/borisveytsman/acmart}
%\acmDataLink{htps://zenodo.org/link}
%%
%% Authors' contribution
%\acmContributions{BT and GKMT designed the study; LT, VB, and AP
%  conducted the experiments, BR, HC, CP and JS analyzed the results,
  %JPK developed analytical predictions, all authors %participated in
  %writing the manuscript.}
%%
%% Sometimes the addresses are too long to fit on the page.  In this
%% case uncomment the lines below and fill them accodingly.
%%
%% \authorsaddresses{Corresponding author: Ben Trovato,
%% \href{mailto:trovato@corporation.com}{trovato@corporation.com};
%% Institute for Clarity in Documentation, P.O. Box 1212, Dublin,
%% Ohio, USA, 43017-6221}
\begin{abstract}

Testing autonomous robotic manipulators is challenging due to the complex software interactions between vision and control components. A crucial element of modern robotic manipulators is the deep learning based object detection model. The creation and assessment of this model requires real world data, which can be hard to label and collect, especially when the hardware setup is not available. The current techniques primarily focus on using synthetic data to train deep neural networks (DDNs) and identifying failures through offline or online simulation-based testing. However, the process of exploiting the identified failures to uncover design flaws early on, and leveraging the optimized DNN within the simulation to accelerate the engineering of the DNN for real-world tasks remains unclear. To address these challenges, we propose the MARTENS (Manipulator Robot Testing and Enhancement in Simulation) framework, which integrates a photorealistic NVIDIA Isaac Sim simulator with evolutionary search to identify critical scenarios aiming at improving the deep learning vision model and uncovering system design flaws. Evaluation of two industrial case studies demonstrated that MARTENS effectively reveals robotic manipulator system failures, detecting 25\% to 50\% more failures with greater diversity compared to random test generation. The model trained and repaired using the MARTENS approach achieved mean average precision (mAP) scores of 0.91 and 0.82 on real-world images with no prior retraining. Further fine-tuning on real-world images for a few epochs (less than 10) increased the mAP to 0.95 and 0.89 for the first and second use cases, respectively. In contrast, a model trained solely on real-world data achieved mAPs of 0.8 and 0.75 for use case 1 and use case 2 after more than 25 epochs.

\end{abstract}


\maketitle

\section{Introduction}

Autonomous robotic manipulators (ARMs), also known as robotic arms, play a crucial role in industrial automation, being widely used for tasks such as palletization and machine tending. To perform tasks autonomously, these systems are equipped with various sensors, including cameras and lidars, allowing them to perceive their surroundings. Data from these sensors is processed by software incorporating machine learning components, such as deep neural networks (DNNs), which excel at detecting visual objects in 2D images, making them suitable for robotic guidance applications.
However, the performance of these models is heavily dependent on the training data, which is typically collected and annotated manually, a process that is labor-intensive. Moreover, the collected samples may not encompass all edge cases that occur in foreseeable operational conditions. One solution to addressing the limited annotated datasets is the creation of synthetic data. This synthetic data can be initially used to evaluate the model's performance beyond the in-distribution samples. A selection of data points can then be incorporated into the next training set to enhance the model's robustness and generalization abilities. 
 
On one hand, when synthetic data is generated through image transformations \cite{pei2017deepxplore, braiek2019deepevolution} or generative models \cite{zhang2018deeproad}, there is no guarantee that it accurately represents naturally occurring use cases. These methods primarily introduce noise and distribution shifts to challenge the model's robustness against alterations in the input data. However, they often fail to produce operational corner cases that reflect specific, adverse environmental conditions. Consequently, while these techniques test the DNN's resilience to input variations, they may not adequately simulate the real-world complexities that the DNN will encounter in practical scenarios. Several methods  \cite{fahmy2023simulator, attaoui2024search} leverage simulation environments to generate challenging and fault-revealing test inputs for the DNNs that are used to improve the DNN model for predictions in the real world.
These works however, only consider the offline test data generation despite numerous studies \cite{haq2020comparing, stocco2023model} highlighting the importance of conducting online testing for autonomous robotic systems with DNN-based components. Indeed, on-line evaluation of DNNs is important as statistical metrics like F1-score and mean Average Precision (mAP) do not allow to capture the failures arising from the DNN interactions with the environment and other system components. %The lack of a closed-loop with the DNN predictions' client component makes it unclear to what extent synthetic data will be useful for real-world tasks. While we typically compute DNN performance using statistical metrics like F1-score and mean Average Precision (mAP), determining the required performance level is challenging. For instance, it is difficult to ascertain how many pixel errors in a segmentation will result in the system's inability to perform tasks such as navigating to a specific position or picking and placing an object. This uncertainty highlights the need for online in-simulation testing methods that consider the actual implications of DNN performance in practical usage scenarios.

Existing simulation-based testing methods are dedicated either to autonomous driving systems (ADS) \cite{klikovits2023frenetic, zohdinasab2021deephyperion, riccio2020model} or unmanned aerial vehicles (UAVs) \cite{khatiri2024sbft}, with few studies focusing on ARM testing \cite{adigun2023risk}. These methods often aim to expose weaknesses in DNNs by generating adversarial test cases through closed loop simulation of the system but do not propose solutions for repairing the identified failures. Therefore, we identify the need for test generation approaches that include both simulation-based online testing of autonomous robotic systems (ARS) with DNN components, as well as the system improvement from failures. Moreover, when improving the perception system from data collected in simulation, it is important to evaluate the usefulness of the improved system on the real world data. %Although some research has been conducted to repair DNNs using simulation-generated data, these studies often fail to evaluate the effectiveness of these repairs in the real-world context, where the DNN is tested with real images. This gap creates a disconnect between the simulation environment and the real-world environment, making the simulation-to-real-world transfer uncertain. Consequently, both testing and repair efforts become focused on simulated problems without providing a systematic engineering pathway for deploying the approved DNN into production.

To address the existing gap in the state-of-the-art approaches for DNN online testing and repair, we propose the MARTENS (Manipulator Robot Testing and Enhancement in Simulation) framework. This framework combines a photorealistic NVIDIA Isaac Sim simulator with evolutionary search for identifying critical scenarios, collecting useful data for system design assessment, testing and improving vision models. Using NVIDIA Isaac Sim, we created a photorealistic simulation environment for ARM applications, enabling the collection of synthetic 2D annotated images for training the DL vision models. The trained DL model is then deployed within the simulation to guide the robot in its manipulation tasks. MARTENS leverages evolutionary search to adjust structural parameters such as object positions and lighting, producing in-simulation tests that reveal failures. These failed test cases provide valuable data for refining the DL model and investigating the limitations of the control and perception strategies. Our evaluation, conducted through two industrial use cases involving pick-and-place tasks, demonstrates that our approach can effectively reveal diverse failures and can be leveraged to repair the DL model. On average, 31\% of the generated in-simulation test cases resulted in the ARM failing the task. By fine-tuning the DL models with synthetic images from failed test cases, we achieved a 99\% success rate in fixing these issues. A detailed analysis of the non-repaired test cases led to the discovery of some important robotic application design flaws. Finally, MARTENS facilitates the convergence to an optimal vision model using synthetic datasets, providing a pre-trained model that engineers can fine-tune with real-world datasets. An effective sim-to-real model transfer procedure reduces the need for extensive real-world data collection and labelling as well as online testing of the vision model using a real robotic system. We evaluate our model trained with synthetic data on the real world images, matching the scene represented in the simulation environment. The model trained and repaired using the MARTENS approach achieved mean average precision (mAP) scores of 0.91 and 0.82 on real-world images with no prior retraining. Further fine-tuning on real-world images for a few epochs (less than 10) increased the mAP to 0.95 and 0.89 for the first and second use cases (UC-1 and UC-2), respectively. In contrast, a model trained solely on real-world data achieved mAPs of 0.8 and 0.75 for UC-1 and UC-2 after more than 25 epochs.
%Further fine-tuning on real-world images for a few epochs (less than 10) increased the mAP to 0.95 and 0.89 for the first and second use cases, respectively. In contrast, a model trained solely on real-world data achieved mAPs of 0.8 and 0.75 after more than 25 epochs.

%The number of training epochs was reduced by 65.3\% and 53.8 \% for use case 1 and use case 2, when using the DNN model pre-trained with MARTENS approach. %....\Houssem{Please continue}
The rest of this paper is structured as follows: Section \ref{sec:approach} describes the details of MARTENS approach for in-simulation DNN testing and improvement. Evaluation results are presented in Section \ref{sec:results}. Section \ref{sec:related_work} discusses the relevant related works, and Section \ref{sec:conclusions} concludes the paper.

\section{Approach}\label{sec:approach}
\begin{figure*}[h!]
\centering
  \includegraphics[scale=0.37]{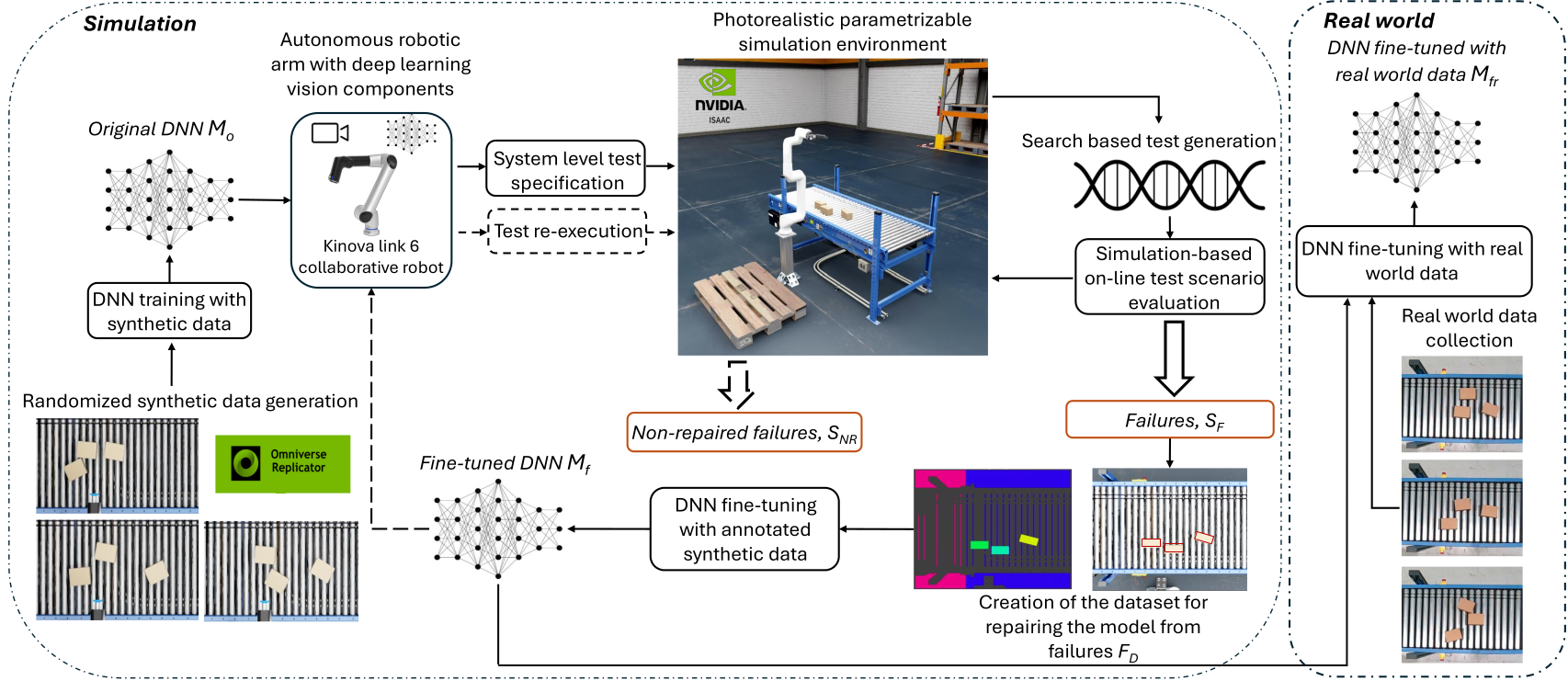}  
\caption{The diagram of the MARTENS approach}
\label{fig:MARTENS}
\end{figure*}

The overall diagram of the MARTENS approach illustrating the major steps is shown in Figure \ref{fig:MARTENS}. In this section, we provide a detailed overview of each of the indicated steps. 

\subsection{In-simulation Test Environment Specification}
The first step in testing perception components in the simulation is setting up a photorealistic parametrizable environment. It should mimic the production environment and produce the variability needed for testing the vision system by randomizing parameters. In a typical environment for testing autonomous robotic manipulators, there are objects with fixed position (static objects) $O_s$, objects with variable position (flexible objects) $O_f$, a robot $R$, a light source $L$ and a camera $C$ with a preset angle of view. To create this typical environment within Isaac Sim, the robot $R$ should be imported using its standard format, URDF (Unified Robot Description Format), which is generally provided by the manufacturer. The robot's URDF contains its physical description, including joints, mass, etc. The objects $O_s$ and $O_f$ can be imported to the simulation from their corresponding 3D computer aided design (CAD) models. The light sources and virtual cameras are included within Isaac Sim and can be conveniently configured. Indeed, Isaac Sim features different types of light sources \cite{nvidia_isaac_sim_lighting} such as rectangular light, distant light, sphere light, cylindrical light, etc. To replicate a conventional lighting condition, we can place the light source $L$ higher in the ceiling and use the rectangular light type that emits a uniform illumination. Varying luminosity is important for testing the robustness of vision models \cite{ebadi2021efficient}. %Figure\ref{fig:light} and Figure\ref{fig:dark} illustrates two different levels of luminosity from a rectangular ceiling light source.

\begin{comment}
\begin{figure}[h!]
\centering
\begin{subfigure}{0.24\textwidth}
  \includegraphics[scale=0.27]{images/uc2_light.PNG}  
  \centering
  \caption{Environment with luminosity set to 4500 }
  \label{fig:light}
\end{subfigure}
\begin{subfigure}{0.22\textwidth}
  \includegraphics[scale=0.27]{images/uc2_dark.PNG}  
  \centering
    \caption{Environment with luminosity set to 1500}
    \label{fig:dark}
\end{subfigure}
\caption{Illustration of the in-simulation test environment}
\label{fig:luminosity}
\end{figure}
\end{comment}
\begin{figure}[H]
\centering
\begin{subfigure}{0.24\textwidth}
  \includegraphics[scale=0.17]{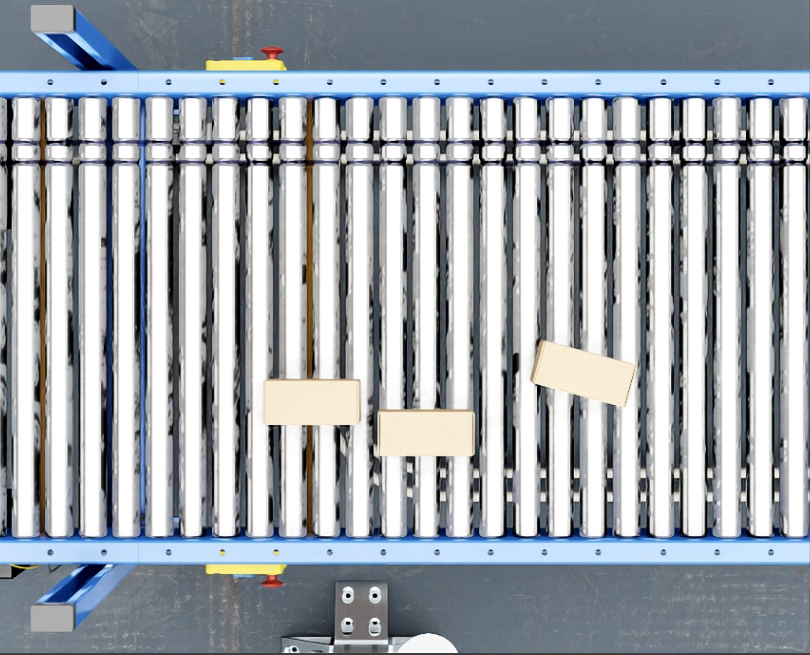}  
  \centering
  \caption{RGB image from the camera}
  \label{fig:img_rgb}
\end{subfigure}
\begin{subfigure}{0.22\textwidth}
  \includegraphics[scale=0.18]{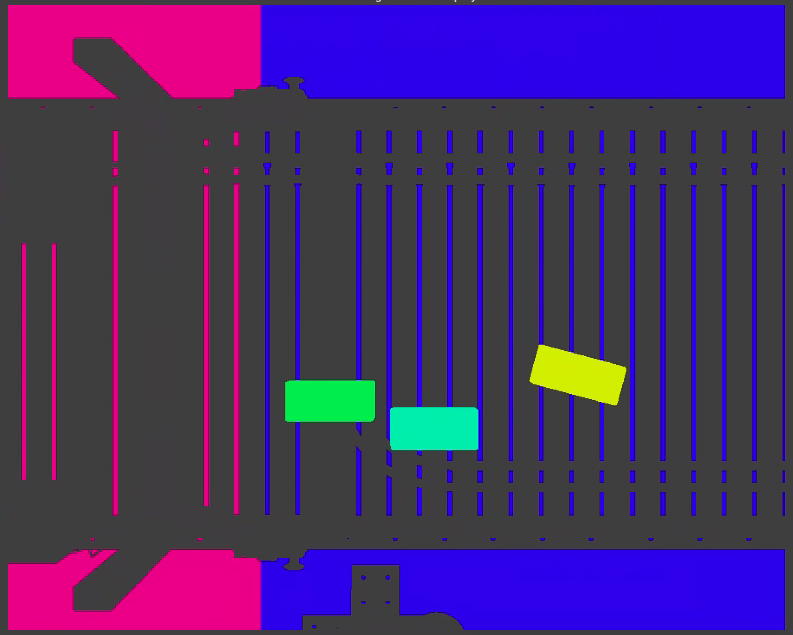}  
  \centering
    \caption{Segmented image }
    \label{fig:img_segmented}
\end{subfigure}
\caption{Camera view in the virtual environment}
\label{fig:segmented}
\end{figure}
The camera $C$ has a pre-defined angle of view and provides RGB images of the virtual environment that are then processed by the vision system. The camera can be located in a fixed position in the scene, also known as `eye-to-hand' camera or it can be mounted on the robot, known as `eye-in-hand' camera. Figure \ref{fig:img_rgb} displays an image captured by an `eye-to-hand' camera mounted on top of the manipulable objects. At inference time, the DL vision predicts the target outputs given the RGB images captured by the virtual camera. However, the estimated pixel coordinates within the 2D images should be projected into the world space of the robot. In real-world production settings, the projection matrix is generally computed using 3D camera calibration \cite{corke2011robotics}. Isaac Sim API provides a function that allows to obtain the coordinates in the world frame from the pixel coordinates in the image. This eliminates the need to calibrate the camera in the simulation and allows for testing the DL vision model with an ideal camera calibration. Moreover, the simulator provides the ground truth information on the position and class of every object in the scene, as shown in Figure \ref{fig:img_segmented}.

The ARM $R$ is tasked with a pick and place task of the flexible objects $O_f$. The ARM should satisfy two requirements while completing the task. The first requirement $R_1$ is that each placed $O_{fi}$ should be placed parallel to the palette with no more than $\phi_d$ degrees of deviation. The second requirement is to perform the placement of each $O_{fi}$ in the target location.
Having set up the test environment in the simulator, we determine the environmental factors that could influence the vision model predictions and the outcomes of the vision-guided robot movements. Given the typical environment of an ARM application described above as well as the system performance requirements, 
we find that the location and orientation of the flexible objects and the lighting intensity affect the vision model predictions. %, resulting in inadequate grasp and placement of the object. 
%\Dima{These parameters can also be identified during the synthetic data generation phase, by evaluating their impact on the model performance.} 
 In order to perform the in-simulation testing, these parameters must be dynamically changed through simulation control logic. In the following, we discuss how these parameters are encoded into a vector to allow searching algorithms to explore the parameter space. The goal of the search is to find such set of parameters that make the ARM falsify the established requirements $R_1$ and/or $R_2$.
\subsection{Search-based Test Generation}
In simulation, randomizing environmental parameters can help generate valid test cases for evaluating vision-guided robotic systems. Nevertheless, the main purpose of testing is to reveal failures by which we can identify design flaws, bugs in the software, and DNN inefficiencies. There are many methods to guide the search of environmental parameters towards the failure-revealing regions. Previously, evolutionary algorithm-based approaches were proven effective at test generation for autonomous robotic systems \cite{gambi2019automatically, birchler2023single}. Below, we provide a more detailed description of the elements of our genetic algorithm (GA) based test generation approach. 

\textit{Individual representation and sampling.} We represent the individuals, or chromosomes, as one-dimensional arrays of size $3n+1$. It contains the concatenation of the $n$ flexible objects' positions, defined as sub-vectors of [$x_i$, $y_i$, $r_i$], where $x_i$, $y_i$ is the object position in the 3D space with the z-coordinate fixed at a pre-defined value and $r_i$ is the rotation angle of the object w.r.t z-axis. The rotation angles w.r.t x and y-axis are fixed and set to 0. The last dimension corresponds to the luminosity $l$ level in the scene. We define constraints to delimit the vector space in order to ensure that: (i) no intersection occurs between the objects; (ii) positions of objects are in the camera's field of view. Indeed, our use cases involve pick and place operations of objects on a planar surface with known height (the z-coordinate is constant or can be read from a depth sensor). Hence, (x, y) coordinates and rotation with respect to the z-axis are sufficient to represent the locations and orientations of objects $O_f$ posed on a planar surface. %\Dima{I think we can remove this text to save space:} In more general use cases, 3D object pose estimation is needed to handle objects that are not confined to a planar surface. This involves determining the object's position in 3D space (x, y, z) and its full orientation (rotation about the x, y, and z axes). This is essential for accurately grasping and manipulating with objects in a three-dimensional environment, where their placement and orientation can vary in all directions.

\textit{Fitness function.} 
Fitness function quantifies how well an individual meets the objective, guiding the selection of superior individuals over the generations \cite{eiben2015introduction}. Robot movement success or failure is a binary information that leads to a sparse fitness function. A continuous score is more suitable for comparing individuals in a generation. Since the DL vision model detects the objects in the 2D image, we calculate the fitness based on the maximum deviation of the estimated object position by the vision system from the actual object position as well as the maximum deviation of the predicted rotation angle from its associated ground truth. The fitness can be formulated as follows:
\begin{align}\label{eq:fitness}
%\[
F &= \frac{w_1}{k_{p}} \cdot \max_{n \in N} \| \ p_n - p_n' \| + \frac{w_2}{k_{r}} \cdot \max_{n \in N} \| r_n - r_n' \|,
%\] 
\end{align}
where $N$ is the total number of objects. Parameter $p_n$ corresponds to the predicted position of the object and $p_n$' of the ground truth object position. $r_n$ corresponds to the estimated object rotation, while the $r_n$' is the actual object rotation. The two deviations w.r.t object location and orientation are not conflicting objectives, so we can use a single-objective genetic algorithm \cite{eiben2015introduction}, rather than a multi-objective one, such as NSGA-II \cite{deb2002fast}. The coefficients $w_1$ and $w_2$ are used to assign weight to objectives. The $k_{p}$ and $k_{r}$ coefficients normalize the values of the objectives to be from the same range. 

\textit{Crossover operator.} We are using a one-point crossover operator, which is one of the commonly-used operators \cite{umbarkar2015crossover}. Given a pair of parents the crossover point is randomly determined. At this point, genes %(their vectors' components) 
(components of the chromosomes) are exchanged. Following the generation of the offspring, a constraint feasibility validation is performed. If the constraint is violated, a low fitness value is assigned to the individual.
%the first segment of the first parent up to the crossover point is combined with the second segment of the second parent to form one offspring, and the reverse combination creates the second offspring. One-point crossover promotes the development of offspring with valuable traits inherited from both parents, increasing the chances of maintaining or improving their fitness. This operator is applied with probability $p_{cross}$. Following the generation of the offspring, a constraint feasibility validation should be performed. For instance, it is important to ensure that the objects combined from the parents do not collide. If the constraint is violated, a very low fitness value is assigned to the individual.

\textit{Mutation operator.} We use two mutation operators: random modification operator $M_{rm}$ and replacement operator $M_r$. $M_{rm}$ operator choose $k$ arbitrary elements of the chromosome (i.e. genes) and then increases or decreases the value of this gene by 10 \%. We vary $k$ from 1 to the chromosome size. $M_r$ operator randomly chooses one object described by a set of $(x_i, y_i, r_i)$ values and replaces them with randomly sampled values. The intuition is to change the position and orientation of one of the objects. At the same time, it is ensured that new object position will not cause collisions with the existing boxes. At each generation, one of the operators $M_r$ or $M_{rm}$ is applied with the probability of $p_{mut}$.

\textit{Duplicate removal.} At each iteration, we remove the individuals which have a small deviation between them. We estimate the deviation as the cosine distance $D_c$ \cite{foreman2014cosine} and compare it to a prefixed minimum threshold, $D_{cth}$. If the $D_c$ between the two individuals is less than $D_{cth}$, one of them is removed. 

The described algorithm is used to generate the test cases that are then evaluated in the simulation environment. Tests are considered failed if the robotic system is unable to pick or place the objects properly. In simulation, the obtained system behaviors can be compared to their desired counterparts, and a maximum deviation threshold can be established to separate passed from failed tests. We define the thresholds to identify the system failures in our use cases in Section \ref{sec:set-up}. As a baseline for the evolutionary search, we implement random search (RS) to generate test cases by arbitrarily selecting values from the allowable ranges. With proper configuration, GA-based search should outperform RS in terms of the proportion of failures revealed while maintaining diversity among the detected failures. For both competing search strategies, the test generation budget is defined by either the allowable total execution time or the maximum number of evaluations. Moreover, the duplicate removal step is applied to both GA and RS.
\subsection{Test Evaluation}\label{sec:test-eval}
Generally, we consider the test failed if at least one of the performance requirements is falsified i.e. the object is not placed correctly on the pallet (i.e., it is not aligned with the pallet, failing the requirement \(R_1\)) or it is not successfully picked and transported to the target position (failing requirement \(R_2\)). During the test generation, both passed and failed test cases are saved in the corresponding sets $S_P$ and $S_F$. As we focus on investigating and repairing the test cases with DNN mispredictions, we further split the failures into two subcategories: (i) \textit{soft} failures, where the DNN prediction was not accurate, leading to a violation of one of the requirements; (ii) \textit{hard} failures, where the DNN prediction was acceptable, but the system failed to satisfy either the requirement $R_1$ or $R_2$.
Afterwards, we analyse on characteristics of the failed tests $S_F$ in terms of structural features sparseness (test diversity) and severity sparseness (failure mode diversity). Indeed, we leverage the failure sparseness, $S_{av}$, a metric proposed in the literature \cite{riccio2020model}, to estimate the diversity by computing the average maximum value of the distance metric between each pair of the failed test cases $tc$:
\begin{equation}\label{eq:spars}
S_{av} = \frac{\sum_{i}^{N} \max_{j}^{N}\text{dist}(tc_i, tc_j)}{N}
\end{equation}
where $N$ = $|S_F|$ is the total number of the failed test cases, and $dist$ is the distance metric. We used the cosine distance metric $D_C$ to compute structural features sparseness, $S_{avf}$.
To evaluate the severity sparseness, we defined five possible failure modes of the system, $FM$, based on domain knowledge and requirements: incorrect box center predicted by DNN $FM_1$, incorrect box orientation predicted by DNN $FM_2$, failure to place the box $FM_3$, incorrect box orientation at placement $FM_4$ and robot stuck during the execution $FM_5$. For each test case, we collect all the failure modes observed during the pick-and-place of each flexible object, $O_f$. Then, to evaluate the diversity of failure modes, we use Eq. (\ref{eq:spars}), but with a different distance metric $dist$. We used the number of the identified unique combinations of the failures modes $N_{FM}$ to estimate the distance between two test cases, thereby computing the severity sparseness, $S_{avs}$. Intuitively, the more unique combination of failure modes are discovered, the better output behavior coverage is achieved.%The more failure modes the test case covers, the more system behavior coverage it achieves.  
\subsection{Vision Model with Randomized Synthetic Data}\label{sec:data_collection}
%The real-world data can be expensive and time-consuming to collect and label manually. The labelling process can be partially automated by integrating pre-trained segmentation models, however, oftentimes these models perform poorly on less common or client-specific objects. At the same time, it is crucial for the perception module development team to be able to test the performance of the developed perception algorithm at the early stage, without waiting for the real-world dataset available. 
As a simulation platform, we selected Nvidia Isaac Sim, which enables advanced GPU-enabled physics simulations with high photorealism. It includes Isaac Replicator \cite{nvidia_omniverse_replicator} that relies on environmental parameters to provide image metadata information such as depth maps and 2D annotations in the form of color-based segmentation (see Figure\ref{fig:img_segmented}). For pick-and-place applications using 2D vision, we post-process the segmented images to extract the contours of the objects of interest $O_f$, then, the polygon coordinates delimiting each object are derived to be added as groundtruth annotations with the RGB images. %The label format is illustrated in Table \ref{tab:obb}, where $x_{j}, y_{j}$ correspond to the coordinates of the polygon of the given object of interest $O_{fi}$.
Using the Isaac Replicator API, we propose generating annotated synthetic images by randomizing environmental parameters within specified distributions. This approach produces scenes with diverse object positions and orientations ($x_i$, $y_i$ and $r_i$), as well as varying luminosity levels $l$. The resulting synthetic dataset enables the vision model to learn and generalize, effectively, improving its ability to accurately detect and position objects under various conditions. Then, after training, the model's performance is evaluated using statistical metrics like mean Average Precision (mAP) based on synthetic validation data collected the same way as the training dataset. The average of mAP is calculated at varying intersection over union (IoU) thresholds, ranging from 0.50 to 0.95 This offline validation process mirrors how vision engineers assess their models' performance to optimize architecture and hyperparameters. However, achieving a perfect mAP of $1.0$ is unlikely to be feasible and can even indicate overfitting. Therefore, testing the vision model in production, integrated with the robotic arm, becomes crucial to verify that the DNN's precision is adequate for ensuring successful picking of the object and placement with correct orientation. In the following sections, we discuss the online evaluation of the ARM with the vision model operating within the above-introduced flexible simulation environment. The proposed in-simulation testing serves to refine the vision model by fine-tuning it on examples with its mispredictions and failures, in addition to uncovering potential design flaws in the vision-based robotic arm system early on. 

\subsection{Model Repair Using Failure Data}\label{sec:repair}
The dataset for repairing the DNN model includes all the RGB images of the scene with objects $O_f$ from the camera $C$, when the ARM failures were detected. For each image, the segmentation data is extracted from the simulator. After post-processing the segmented images, we obtain a dataset $D_F$ of images and labels suitable for model training. This dataset includes all failures identified by the generated test cases, $S_F$, along with a selection of passed tests from $S_P$, where the system completed the task but the DNN predictions were inaccurate compared to the ground truth provided by Isaac Sim. Specifically, these are cases where the predicted orientation of the cardboard boxes deviated from the ground truth by more than 5 degrees, or where the predicted center of the boxes deviated by more than 1 cm. These test cases can be categorized as near-fails, denoted $S_{NF}$. Adding them to our training dataset for fine-tuning can help the model improve its inductive bias and enhance its performance, reducing the likelihood of such cases turning into actual failures.
%\Dima{Our results should be replicable, we need to define what is $S_{NF}$!} \Dima{Test case is categorized as $S_{NF}$ is the predicted box center deviates by more than 2 cm from the ground truth or if the detected orientation deviates more than 5 degrees from the ground truth}. 
Thanks to transfer learning, the DL vision model does not need to be retrained from scratch on the entire dataset, i.e., combining the original and the novel datasets. Instead, the fine-tuning can be done directly on the last operating model, $M_o$, considering it as the base (pretrained) model, and fine-tuning it exclusively on the failed dataset, $D_F$, using the same hyperparameters, except for decreasing the maximum number of epochs. This reduction helps mitigating the risk of catastrophic forgetting phenomenon, where the model overfits to the new samples to the extent of losing its original performance. Furthermore, we use a merged validation dataset that includes original validation data and novel validation data (the collected failed test data). Even with high performance on validation set, offline assessment alone cannot ensure that most of the failures are resolved. It is essential to replay the revealed failures in a subsequent in-simulation testing session using the fine-tuned model $M_f$. Thus, we re-run the failed test cases from $S_F$ and record a new set of test cases that passed after the model repair $S_R$ (`repaired' test cases) and a set of test cases that did not pass $S_{NR}$ after the repair (`non-repaired' test cases). As a result, we will be able to measure how many failures were fixed by DL vision model repair.
%\Dima{We don't need this sentence, why are you talking about multiple repair cycles?}This comprehensive assessment ensures that the model adequately fixes its erroneous predictions while still handling past data, contributing to sustained performance over multiple repair cycles. \Dima{I don't understand, you say we merged them, now you say only about failed failed dataset. Let's remove the next sentence}. It is worth noting that the new validation dataset was derived by partitioning the failed dataset according to the same training/validation split ratios utilized in the initial training phase. 

\section{Evaluation}

\subsection{Experimental Setup}\label{sec:set-up}
\begin{figure}[h!]
\centering
\begin{subfigure}{0.22\textwidth}
  \includegraphics[scale=0.25]{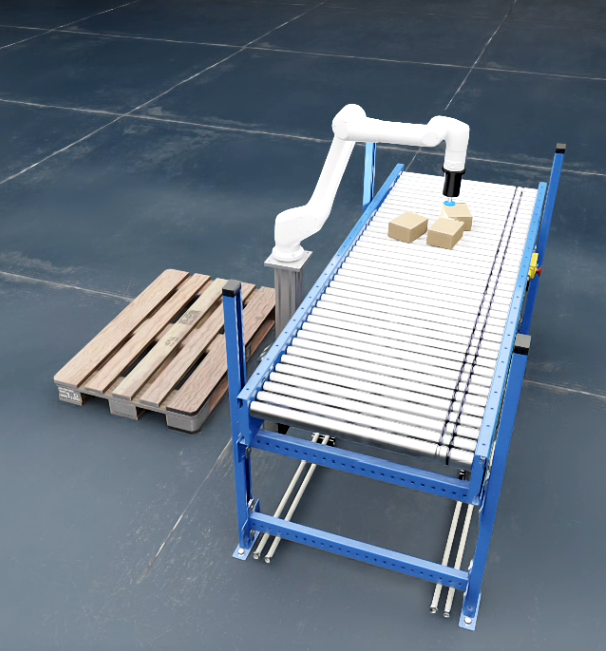}  
  \centering
  \caption{Manipulator with suction gripper }
\end{subfigure}
\begin{subfigure}{0.22\textwidth}
  \includegraphics[scale=0.23]{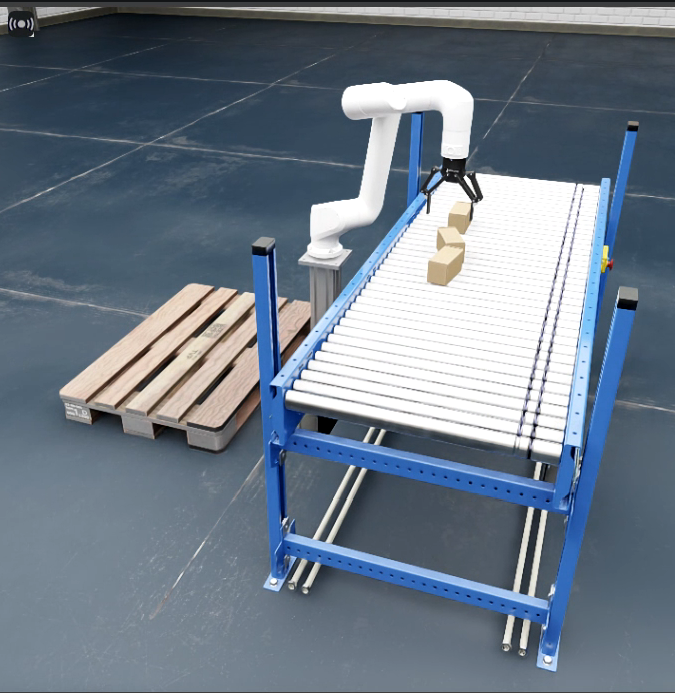}  
  \centering
    \caption{Manipulator with parallel gripper}
\end{subfigure}
\caption{Illustrations of the test subjects used in the study}
\label{fig:robots}
\end{figure}

\textbf{Use cases}. We evaluate our approach on two test subjects that are autonomous robotic manipulators performing pick and place task as illustrated in Figure \ref{fig:robots}. The task aims to automate the process of pulling cardboxes from a conveyor belt and arranging them appropriately on a pallet for delivery. In this task, the robotic manipulators are used for streamlining packaging operations and ensuring efficient transport loading. For both test subjects we use a Kinova link 6 \cite{kinova} collaborative robot. First test subject (test subject 1) is equipped with suction gripper, while the second subject (test subject 2) with a parallel gripper (Robotiq 2F-85 \cite{robotiq_gripper}). Test subject 1 tackles the task of pick and place cardboxes of the 17 x 14 cm size, while the test subject 2 handles the 12 x 8 cm cardboxes. Parallel grippers typically require more precise positioning and control to ensure a successful grasp, than the suction grippers \cite{monkman2007robot}. Both systems have a fixed camera mounted on top of the conveyor belt's region of interest, i.e., reachable by the robot. The camera is an Intel RealSense D435 \cite{intel_realsense_d435}, configured to capture images of the size 640 x 514 pixels.

The robot needs accurate vision guidance to achieve the grasping of the objects and then complete their proper placement in the pre-defined locations (i.e., properly placed means correct position and orientation). %Figure \ref{fig:diagram} represents the controller state diagram. 
The robot starts in the default position, also known as `home' position. Then, the image is captured with the fixed camera. If cardboxes are detected by the DL model in the image, the controller initiates the pick and place cycle of the box with the highest detection confidence. The box center coordinates are translated into the robot's coordinate frame and the rotation angle w.r.t the z-axis is sent to the robot to grasp the cardbox with the correct orientation. Once the box is securely grasped, the robot lifts it off the conveyor belt and transports it to the place position. Then, the gripper is opened, box is placed and the robot returns to the home position, where the cycle repeats. When no cardboxes are detected, the controller is stopped. Typically, this cycle is repeated for 3 times as we fixed the number of cadboxes to 3. To implement robot control within Isaac Sim, we used the RMPFlow controller \cite{cheng2020rmp}, which encapsulates an inverse kinematics solver and enables the implementation of robot primitives for the necessary movements based on computed and vision-estimated positions. When the actual robotic system is implemented, these primitives will be handled by Kinova's Kortex API \cite{kinova_kortex_api}. 
In both use cases, two operational requirements are defined as follows: \(R_1\) - place the object with a 0-degree orientation relative to the pallet  and \(R_2\) - perform the pick and transport the object to the target position. The failure is detected when the box orientation at the place location deviates by more than 5 degrees from the target. Failure is also detected if the box is positioned more than 50\% of its size away from the target location.
\begin{comment}
\begin{figure}[h!]
\centering
  \includegraphics[scale=0.5]{images/diagram2.PNG}  
  \centering
 \caption{State diagram of the pick and place controller for the use cases}
\label{fig:diagram}
\end{figure}
\end{comment}
\begin{figure}[h]
\centering
\begin{subfigure}{0.2\textwidth}
  \includegraphics[scale=0.1]{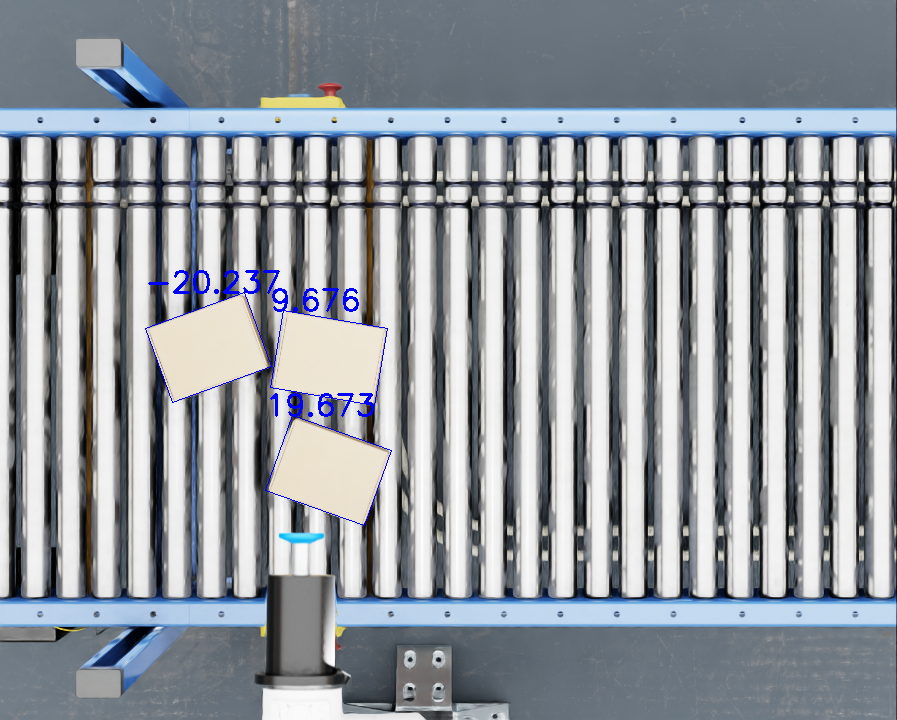}  
  \centering
  \caption{Test subject 1 view}
\end{subfigure}
\begin{subfigure}{0.2\textwidth}
  \includegraphics[scale=0.1]{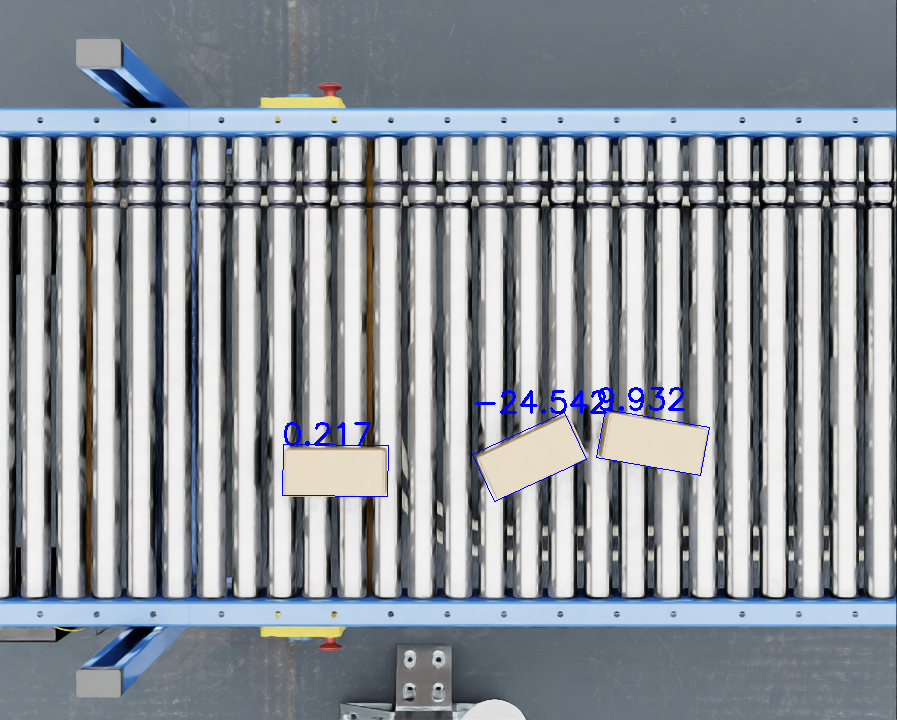}  
  \centering
  \caption{Test subject 2 view}
\end{subfigure}
\caption{Top views from the fixed camera for the subjects under test with model predictions}
\label{fig:camera_view}
\end{figure}

\textbf{Vision models.} The vision models are responsible for detecting the positions and orientations of the target objects i.e. cardboxes. An example of the model output predictions are shown in Figure \ref{fig:camera_view}. Indeed, the vision system should send the coordinates of the cardbox center and its rotation angle w.r.t the z-axis. This information can be inferred from the 2D images using YOLOv8 \cite{yolo_v8} that natively supports the oriented bounding box detection. As explained in Section \ref{sec:data_collection}, images and coordinates of polygons defining cardboxes will be prepared. Such labelled datasets can be used to train YOLOv8 for oriented bounding box detection \cite{oriented_bounding_boxes}: its center location, height, width as well as its orientation. For each of the use cases, we trained a dedicated vision DL model using 1200 synthetic images. The images were collected by randomizing the scene parameters through the Isaac Replicator API following the uniform sampling from the ranges shown in Table \ref{tab:repr} with the orientation of the boxes ranging from -25 to 25 degrees. 
We then split the dataset into train set of 960 images and a validation set of 240 images. As recommended by Yolov8 creators \cite{yolov_training}, we used a batch size of 16, an adaptive learning rate from 0.01 to 0.0001, and image augmentations like changing the brightness, rotation, scaling, flipping and shear. Using early-stopping on validation data, we trained the models for 100 epochs and both best-fitted models achieved a mean average precision (mAP) of 0.98. Having such high precision is expected since the synthetic dataset is collected using a simulation environment and a uniform sampled data from the same distribution. Online in-simulation testing of the model aims to find inputs that reduce its precision in identifying objects in the scene.
%\Dima{ When the model is tested online,  we aim to find such inputs, that reduce the precision of the model in the identification of the objects in the scene.}  \Dima{remove this} precision can be reduced due to potential edge cases. \Dima{Let's remove this sentence because we do not address this issue}. However, the goal is to identify the number of failures, as the required level of vision model precision is determined by successfully passing the pick-and-place test cases. The requirements for a vision-based robotic arm are typically formulated in terms of reliability and robustness in performing its primary tasks.

To run the experiments we used `g5.2xlarge' \cite{aws_ec2_g5} AWS instance with the following characteristics:  A10G Tensor Core GPU with 24 Gb of memory, 8 AMD EPYC CPUs, 32 Gb RAM. In comparing the results obtained from the randomized algorithms, we repeated all evaluations at least 5 times due to the computational cost of running the simulations. For all results, we report the non-parametric %Mann-Whitney U p
permutation test\cite{good2013permutation} with a significance level of $\alpha = 0.05$, as well as the effect size measure in terms of Cliff's delta \cite{macbeth2011cliff}. Given the computational constraints to obtain more runs of the algorithms, we chose permutation test as it makes fewer assumptions about the input data compared to other non-parametric tests such as Mann-Whitney U test~\cite{mann1947test}.

\textbf{Search algorithm configuration.} 
For both use cases, we fixed the number of flexible objects $O_f$, i.e. cardboxes, to $N=3$. The chromosomes are 10-D arrays with the first 9 dimensions representing the positions ($x_i$, $y_i$) and rotations ($r_i$) around the z-axis of the three cardboxes. The last dimension represents the luminosity level $l$ at the scene. The initial population is generated by randomly sampling the values in the allowed ranges, as specified in Table \ref{tab:repr}. 
\begin{table}[h!]
    \caption{Allowed ranges for the parameters}
    \centering
    \scalebox{0.9}{
    \begin{tabular}{ccccccc}
    \hline
       $x_{i}$ & $y_{i}$ &  $r_{i}$ & $l$  \\
    \hline
       $0.5 - 0.9 $ & $0 - 0.92$ & $-30 - 30$ & $1500-5000$   \\
      \hline
    \end{tabular}}

    \label{tab:repr}
\end{table}
We calculated the fitness function using Eq.~(\ref{eq:fitness}), with equal weight coefficients $w_1$ and $w_2$ set to 0.5. The normalization coefficients $k_p$ and $k_r$ were chosen based on previously recorded values and set to $0.01$ and $1$, respectively. We fix the value of mutation probability $p_{mut}$  as $0.4$, crossover rate $p_{cross}$ as $0.9$ and the duplicate removal threshold $D_{cth}$ as $0.1$. These values are influenced by our previous studies \cite{humeniuk2022search} and confirmed in the warm-up experiments. Population size was set to 40 with the execution budget of 220 evaluations. Over the generations, each individual is used to create the test case scene and evaluate the simulated robotic arm in performing the pick and place of the three cardboxes. The evaluation of each test scene using Isaac Sim is computationally expensive and takes on average 280 seconds for use case 1 (UC-1) and 310 seconds for use case 2 (UC-2) using the GPU-powered AWS instance: `g5.2xlarge'. For failure modes, they can be identified as follows: $FM_1$ occurs when the center $O_f$ predicted by the DNN model deviates from the ground truth by more than 1 cm. $FM_2$ is noted when the predicted orientation of $O_f$ deviates by more than 5 degrees from the ground truth. $FM_3$ indicates that $O_f$ was not placed in the target location. $FM_4$ occurs when the orientation of the placed $O_f$ deviates by more than 5 degrees from the target orientation. $FM_5$ denotes when the $ARM$ gets stuck during execution and restarts the operation cycle.\\ %In regards to near-fail tests, $S_{NF}$, they include the incorrect predictions despite the robot's success, i.e., the predicted box center or orientation that deviates by more than 2 cm or 5 degrees, respectively, from the ground truth. 
\textbf{Expert feedback.} Given that this research was conducted in partnership with industrial collaborators from Sycodal, we had the opportunity to gather domain expert feedback on each of our research questions (RQs). We conducted two 1-hour interviews with two robotic and vision engineers at Sycodal. During these interviews, we first presented the results obtained for each RQ, followed by a 15-minute discussion. We noted all the points mentioned by the engineers and present them at the end of each RQ.
\subsection{Research Questions}\label{sec:results}

\subsubsection{RQ1: Effectiveness of MARTENS in revealing ARM failures} 
%(Usefulness of MARTENS for simulator-based testing of autonomous robotic manipulators). To what extent can MARTENS improve the identification of diverse failures of ARMs compared to baseline approaches?
\hfill \break
\textbf{Motivation.} We aim to ensure that the designed in-simulation testing method is able to reveal system failures. Our goal is also to determine whether evolutionary search (GA) can reveal more diverse and numerous failures than random search (RS).\\
\textbf{Method.} For both use cases, we perform five runs of RS and GA with 220 evaluations each. Then, we compute the number of passed versus failed test cases, as well as the near-failed test cases. To distinguish the failure involving DNN prediction inaccuracy and the system failures even with correct predictions from DNN, we calculate the number of soft versus hard failures. Understanding the failures that occurred when the DNN predictions were adequate is critical as it can point towards some design flows on the control algorithm level. To compare between the GA and RS beyond the number of revealed failures, we measure the structural features sparseness among test environments generated by each method, as well as, the severity sparseness. This gives us the distinctive characteristics among the revealed failures in terms of diversity and coverage. 
\begin{comment}
\begin{table}[!h]
    \centering
    \caption{Generated tests statistics for UC-1}
    \scalebox{0.7}{
\begin{tabular}{lcccccccccc}
        \toprule
        \textbf{Algorithm} & \multicolumn{3}{c}{\textbf{All Tests}} & \multicolumn{3}{c}{\textbf{Failing Requirements}} & \multicolumn{2}{c}{\textbf{Failure Type}} & \textbf{Total} \\ 
        \cmidrule(lr){2-4} \cmidrule(lr){5-7} \cmidrule(lr){8-9}
        & \textbf{Pass} & \textbf{Fail} & \textbf{DNN Fail} & \textbf{r1} & \textbf{r2} & \textbf{r1\&r2} & \textbf{Soft} & \textbf{Hard} & \\ 
        \midrule
        \textbf{RS} & 531 & 532 & 26 & 530 & 0 & 2 & 510 & 22 & 1093 \\ 
        \textbf{GA} & 243 & 794 & 45 & 786 & 0 & 8 & 782 & 12 & 1082 \\
        \textbf{Total} & 774 & 1326 & 71 & 1316 & 0 & 10 & 1292 & 34 & 2175 \\
        \bottomrule
    \end{tabular}}
    \label{tab:test_analysis}
\end{table}
\end{comment}
\begin{table}[!h]
    \centering
    \caption{Generated tests statistics for UC-1}
    \scalebox{0.8}{
    \begin{tabular}{lcccccc}
        \toprule
        \textbf{Algorithm} & \multicolumn{3}{c}{\textbf{All Tests}} & \multicolumn{2}{c}{\textbf{Failure Type}} & \textbf{Total} \\ 
        \cmidrule(lr){2-4} \cmidrule(lr){5-6}
        & \textbf{Pass} & \textbf{Fail} & \textbf{Near Fail} & \textbf{Soft} & \textbf{Hard} & \\ 
        \midrule
        \textbf{RS} & 531 & 532 & 26 & 510 & 22 & 1093 \\ 
        \textbf{GA} & 243 & 794 & 45 & 782 & 12 & 1082 \\
        \textbf{Total} & 774 & 1326 & 71 & 1292 & 34 & 2175 \\
        \bottomrule
    \end{tabular}}
    \label{tab:test_analysis}
\end{table}

\begin{table}[!h]
    \centering
    \caption{Generated test statistics for UC-2}
    \scalebox{0.8}{
    \begin{tabular}{lcccccc}
        \toprule
        \textbf{Algorithm} & \multicolumn{3}{c}{\textbf{All Tests}} & \multicolumn{2}{c}{\textbf{Failure Type}} & \textbf{Total} \\ 
        \cmidrule(lr){2-4} \cmidrule(lr){5-6}
        & \textbf{Pass} & \textbf{Fail} & \textbf{Near Fail} & \textbf{Soft} & \textbf{Hard} & \\ 
        \midrule
        \textbf{RS} & 1038 & 27 & 33 & 9 & 18 & 1100 \\ 
        \textbf{GA} & 935  & 34 & 112 & 12 & 22 & 1084 \\ 
        \textbf{Total} & 1973 & 61 & 145 & 21 & 40 & 2184 \\ 
        \bottomrule
    \end{tabular}}
    \label{tab:test_analysis_uc2}
\end{table}

\begin{figure}[h!]
\centering
\begin{subfigure}{0.22\textwidth}
  \includegraphics[scale=0.25]{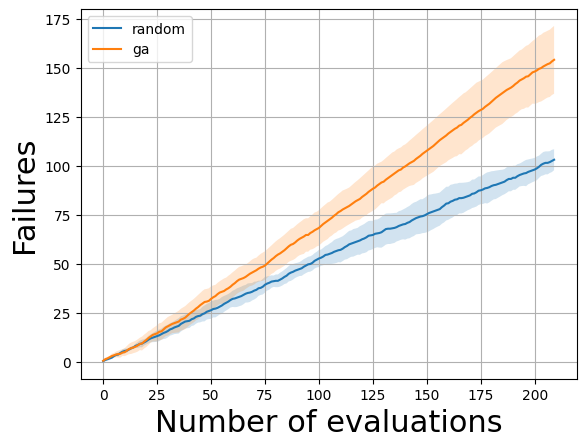}
  \centering
  \caption{Failure convergence for use case 1}
\end{subfigure}
\begin{subfigure}{0.22\textwidth}
  \includegraphics[scale=0.25]{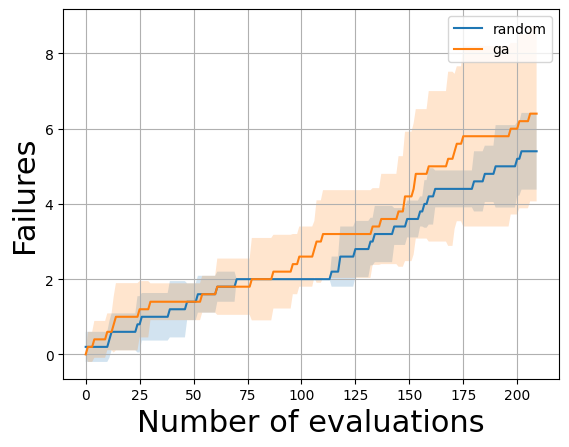}
  \centering
  \caption{Failure convergence for use case 2}
\end{subfigure}
\caption{Number of failures revealed over time}
\label{fig:failure_convergence}
\end{figure}

\begin{comment}
\begin{table}[!h]
    \centering
    \caption{Generated test statistics for UC-2}
    \scalebox{0.7}{
\begin{tabular}{lcccccccccc}
        \toprule
        \textbf{Algorithm} & \multicolumn{3}{c}{\textbf{All Tests}} & \multicolumn{3}{c}{\textbf{Failing Requirements}} & \multicolumn{2}{c}{\textbf{Failure Type}} & \textbf{Total} \\ 
        \cmidrule(lr){2-4} \cmidrule(lr){5-7} \cmidrule(lr){8-9}
        & \textbf{Pass} & \textbf{Fail} & \textbf{DNN Fail} & \textbf{r1} & \textbf{r2} & \textbf{r1\&r2} & \textbf{Soft} & \textbf{Hard} & \\ 
        \midrule
        \textbf{RS} & 1038 & 27 & 33    & 19 & 8 & 0 & 9 & 18 & 1100 \\ 
        \textbf{GA} & 935  & 34 & 112   & 26 & 8 & 0 & 12 & 22 & 1084 \\ 
     \textbf{Total}& 1973 & 61  & 145   & 45 & 16 & 0 & 21 & 40 & 2184 \\ 
        \bottomrule
    \end{tabular}}
    \label{tab:test_analysis_uc2}
\end{table}
\end{comment}
\textbf{Results.} Table \ref{tab:test_analysis} and Table \ref{tab:test_analysis_uc2} present all the statistics of passed and failed test cases, as well as their sub-categories for UC-1 and UC-2, respectively. Overall, our approach reveals 1326 and 61 failures in UC-1 and UC-2, respectively, based on more than 2175 generated test cases. Examples of the revealed failing test cases with DNN mispredictions are available online \cite{humeniuk_2024_12748885}. For the UC-1, the majority of the failures occurred due to model inefficiencies, which affects the precision of vision-based picks and results in 97.4\% soft failures. For the UC-2, 65.5 \% of the failures are hard, meaning they happened even when the DNN model predicted correctly, suggesting that the pose estimation accuracy needs to be increased or potential design flaws exist. 
When comparing the numbers obtained by using each searching strategy, we find that GA-driven test generation reveals 50 \% (UC-1) and 26 \% (UC-2) more failures, than RS-driven test generation. Interestingly, in UC-1 RS reveals more hard failures than RS. We attribute it to the fact that the fitness function is designed to directly promote DNN mispredictions i.e soft failures, rather than the task level failures. Our GA implementation successfully evolves the generations towards being more likely to exploit system vulnerabilities and cause system failures. Figure \ref{fig:failure_convergence} illustrates the evolution of the number of failures revealed by each strategy over the iterations of the search. Initially, the GA-based test generation begins with random test cases, displaying a similar likelihood of inducing failures as the RS approach. However, over successive iterations, the GA method increasingly exposes more failures compared to RS. This widening gap demonstrates that the GA-based generations are progressively improving. In the following analysis, we compare the structural features and severity sparseness among the test cases generated by GA and RS to evaluate them in terms of test diversity and failure mode coverage.

\begin{figure}[h!]
\centering
\begin{subfigure}{0.22\textwidth}
  \includegraphics[scale=0.27]{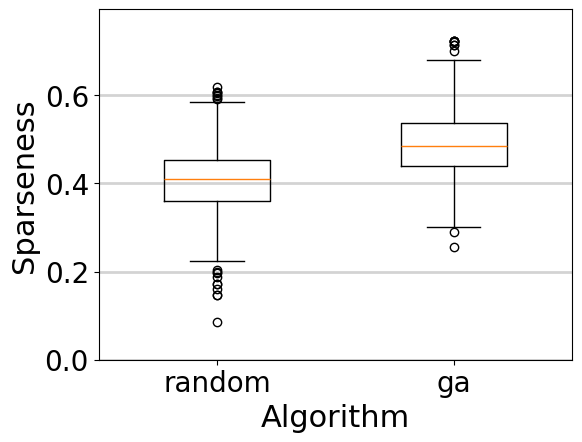}  
  \centering
  \caption{Use case 1 feature sparseness}
\end{subfigure}
\begin{subfigure}{0.22\textwidth}
  \includegraphics[scale=0.27]{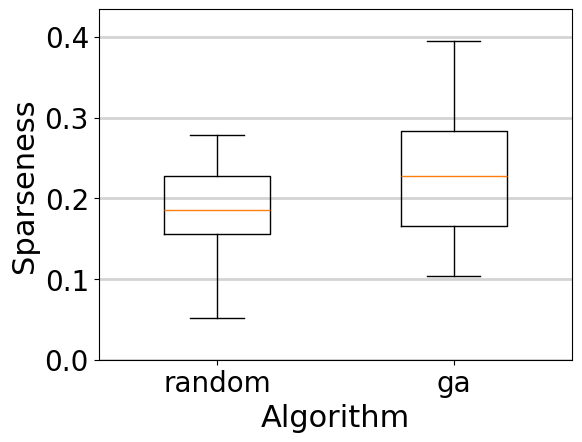}
  \centering
  \caption{Use case 2 feature sparseness}
\end{subfigure}
\caption{Sparseness of the structural features of the revealed failures  for use cases 1 and 2}
\label{fig:input_sparseness}
\end{figure}

Table \ref{tab:sparseness_metrics} presents the sparseness metrics obtained by each search strategy, along with the results of the statistical significance tests comparing their differences. As can be seen, GA generates significantly more diverse test cases in terms of structural features compared to RS, with a large effect size. This can also be observed in Figure \ref{fig:input_sparseness}, which displays box plots of feature sparseness values for test cases generated by GA versus RS. Thus, GA not only increases the number of revealed failures by exploiting the characteristics of previous test cases but also enhances the diversity of features among these test cases. The randomness introduced through crossover and mutation in our implementation promotes novelty in new generations, even though they are derived from previous test cases. In terms of severity sparseness, GA produces higher values than RS; however, there is no statistically significant difference. This indicates that both search strategies were effective in uncovering a variety of failure modes.% and increasing the coverage of system behaviors exhibited in the respective test cases. %The number of failure modes covered during evaluations by each search strategy as shown in Figure \ref{fig:failure_code_convergence}.  %Since GA generated more diverse test cases, it has a relatively higher opportunity for coverage. This is better illustrated by the failure modes covered during evaluations by each search strategy for UC-1, as shown in Figure \ref{fig:failure_code_convergence}. 

\begin{table}[h]
\centering
\caption{Sparseness metrics obtained by each search strategy}
\scalebox{0.8}{
\begin{tabular}{p{1.2cm} p{2.5cm} p{1cm} p{1cm} p{1.2cm} p{1.5cm}}
\toprule
\textbf{Use case} & \textbf{Metric} & \textbf{Random} & \textbf{GA} &\textbf{p-value} & \textbf{Effect size} \\
\midrule
\textbf{uc1} & Features Sparseness & 0.406 & 0.488 & < 0.001 & -0.563 L \\
    & Severity Sparseness & 7.6 & 7.8 & 1 & -0.12 N \\
\midrule
\textbf{uc2} & Features Sparseness & 0.188 & 0.235 & 0.023 & -0.342 M \\
    & Severity Sparseness & 2.4 & 2.8 & 0.683 & -0.28 S \\
\bottomrule
\end{tabular}}
\label{tab:sparseness_metrics}
\end{table}

\begin{comment}
\begin{figure}[h!]
\centering
\begin{subfigure}{0.22\textwidth}
  \includegraphics[scale=0.25]{images/uc1_coverage_convergence.png}
  \centering
  \caption{Use case 1}
\end{subfigure}
\begin{subfigure}{0.22\textwidth}
  \includegraphics[scale=0.25]{images/uc2_coverage_convergence.png}
  \centering
  \caption{Use case 2}
\end{subfigure}
\caption{Number of unique failure modes discovered }
\label{fig:failure_code_convergence}
\end{figure}
\end{comment}
\textbf{Expert feedback.} The design of the robot tool and the pick-and-place movements significantly influence the vision system requirements. For the suction gripper use case, containing a big number of soft failures, the precise position and orientation detection play an important role in the task achievement, as most of the task failures occurred following the mispredictions of the DNN model. At the same time, the ARM with parallel gripper in UC-2 is more prone to hard failures. This is because the gripper, being relatively large (when opened), may collide with other objects during the picking or accidentally contact the other cardbox during movement, causing slight deviations from the intended placement position and augmenting the initial errors from the perception module. To compensate for the influence of external factors, the requirements for object detection precision by DNN should be increased.  %  The precision of the estimated pose required for these movements is task-dependent. Accordingly, the selection of camera resolution, the deep learning model, and the image features to be detected are all driven by the precision needs of the task. For instance, in UC-2, which employs a parallel gripper, there is a higher demand for precise contour detection compared to UC-1, which uses a suction gripper that only requires accurate detection of the cardbox center. 
%Additionally, the parallel gripper in UC-2 is more prone to hard failures. This is because the gripper, being relatively large (when opened), may collide with other objects during the picking or inadvertently contact the cardbox during movement, causing slight deviations from the intended placement position. UC-2 may have other soft failures hidden within the hard ones because we have the same precision requirement for both use cases for vision detection. For instance, the incorrect rotation at placement, can be caused by an increased, but acceptable, deviation of the DNN prediction from the ground truth.

\subsubsection{RQ2: Usefulness of MARTENS in repairing DNN's inefficiencies}
%(Effectiveness of MARTENS for the failure repair). How effective is DNN model fine-tuning in repairing the discovered failures?
\hfill \break
\textbf{Motivation.} We aim to assess the effectiveness of our approach in revealing failures that can be exploited to address inefficiencies in the DNN. We also seek to examine the persistent failures following the DNN repair to understand its limitations and identify potential design flaws.

\textbf{Method.} Based on the obtained failure and near-failures sets $S_F$ and $S_{NF}$, the DL models are fine-tuned for each use case as described in Section \ref{sec:repair}. To compare the usefulness of our GA implementation with simple RS, we create separate fine-tuned models, $M_{f}^{RS}$ and $M_{f}^{GA}$, which are trained exclusively on datasets collected during the execution of tests generated with their respective search strategies, $RS$ and $GA$. Then, we re-run the test cases in the simulation using $M_{f}^{RS}$ and $M_{f}^{GA}$ for each use case. %We measure statistics to evaluate the effectiveness of fine-tuning in addressing these inefficiencies and to compare the two alternative search strategies. 
We calculate the number of passing, failing, and near-failing tests, noting the type of failures: soft or hard. We refer to the failed tests as 'non-repaired' tests, indicating that the failure persists even with the fine-tuned DNN. To account for the non-determinism of the simulator and ensure these failures are not due to simulator flakiness, we re-run each non-repaired test five times.

\textbf{Results.} Tables \ref{tab:test_analysis_repair_uc1} and \ref{tab:test_analysis_repair_uc2} present the statistics on the evaluation of the models fine-tuned with datasets generated by either GA or RS for both use cases.

\begin{table}[!ht]
    \centering
    \caption{UC-1 test analysis after model repair}
    \scalebox{0.8}{
    \begin{tabular}{lcccccc}
        \toprule
        \textbf{Algorithm} & \multicolumn{3}{c}{\textbf{All Tests}} & \multicolumn{2}{c}{\textbf{Failure Type}} & \textbf{Total} \\ 
        \cmidrule(lr){2-4} \cmidrule(lr){5-6}
        & \textbf{Pass} & \textbf{Fail} & \textbf{Near Fail} & \textbf{Soft} & \textbf{Hard} & \\ 
        \midrule
        \textbf{RS} & 557 & 1 & 0 & 0 & 1 & 558 \\ 
        \textbf{GA} & 838 & 1 & 0 & 0 & 1 & 839 \\
        \textbf{Total} & 1395 & 2 & 0 & 0 & 1 & 1397 \\
        \bottomrule
    \end{tabular}}
    \label{tab:test_analysis_repair_uc1}
\end{table}

\begin{comment}
\begin{table}[!ht]
    \centering
    \caption{UC-1 test analysis after model repair}
    \scalebox{0.7}{
\begin{tabular}{lcccccccccc}
        \toprule
        \textbf{Algorithm} & \multicolumn{3}{c}{\textbf{All Tests}} & \multicolumn{3}{c}{\textbf{Failing Requirements}} & \multicolumn{2}{c}{\textbf{Failure Type}} & \textbf{Total} \\ 
        \cmidrule(lr){2-4} \cmidrule(lr){5-7} \cmidrule(lr){8-9}
        & \textbf{Pass} & \textbf{Fail} & \textbf{DNN Fail} & \textbf{r1} & \textbf{r2} & \textbf{r1\&r2} & \textbf{Soft} & \textbf{Hard} & \\ 
        \midrule
        \textbf{RS} & 557 & 1 & 0 & 1 & 0 & 0 & 0 & 1 & 558 \\ 
        \textbf{GA} & 838 & 1 & 0 & 1 & 0 & 0 & 0 & 1 & 839 \\
     \textbf{Total} & 1395& 2 & 0 & 2 & 0 & 0 & 0 & 1 & 1397 \\
        \bottomrule
    \end{tabular}}
    \label{tab:test_analysis_repair_uc1}
\end{table}
\end{comment}
\begin{table}[!ht]
    \centering
    \caption{UC-2 test analysis after model repair}
    \scalebox{0.8}{
    \begin{tabular}{lcccccc}
        \toprule
        \textbf{Algorithm} & \multicolumn{3}{c}{\textbf{All Tests}} & \multicolumn{2}{c}{\textbf{Failure Type}} & \textbf{Total} \\ 
        \cmidrule(lr){2-4} \cmidrule(lr){5-6}
        & \textbf{Pass} & \textbf{Fail} & \textbf{Near Fail} & \textbf{Soft} & \textbf{Hard} & \\ 
        \midrule
        \textbf{RS} & 59 & 0 & 0 & 0 & 0 & 59 \\ 
        \textbf{GA} & 145 & 2 & 0 & 0 & 2 & 147 \\
        \textbf{Total} & 204 & 2 & 0 & 0 & 2 & 206 \\
        \bottomrule
    \end{tabular}}
    \label{tab:test_analysis_repair_uc2}
\end{table}

\begin{comment}
\begin{table}[!ht]
    \centering
    \caption{UC-1 test analysis after model repair}
    \scalebox{0.7}{
\begin{tabular}{lcccccccccc}
        \toprule
        \textbf{Algorithm} & \multicolumn{3}{c}{\textbf{All Tests}} & \multicolumn{3}{c}{\textbf{Failing Requirements}} & \multicolumn{2}{c}{\textbf{Failure Type}} & \textbf{Total} \\ 
        \cmidrule(lr){2-4} \cmidrule(lr){5-7} \cmidrule(lr){8-9}
        & \textbf{Pass} & \textbf{Fail} & \textbf{DNN Fail} & \textbf{r1} & \textbf{r2} & \textbf{r1\&r2} & \textbf{Soft} & \textbf{Hard} & \\ 
        \midrule
        \textbf{RS} & 59 & 0& 1 & 0 & 0 & 0 & 0 & 0 & 60 \\ 
        \textbf{GA} &145 & 2 & 0 & 2 & 0 & 0 & 0 & 2 & 146 \\
     \textbf{Total} & 204& 2 & 1 & 2 & 0 & 0 & 0 & 2 & 206 \\
        \bottomrule
    \end{tabular}}
    \label{tab:test_analysis_repair_uc2}
\end{table}
\end{comment}

\begin{figure}[h!]
\centering
\begin{subfigure}{0.2\textwidth}

  \includegraphics[scale=0.3]{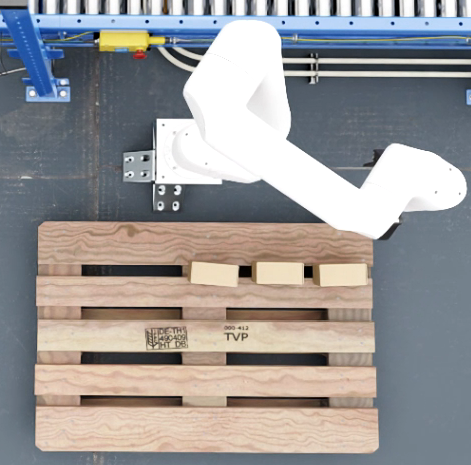}
  \centering
  \caption{Robot misplaces the box after shaking.}
  \label{fig:non_repair_examples_uc2_2}
\end{subfigure}
\begin{subfigure}{0.2\textwidth}
  \includegraphics[scale=0.16]{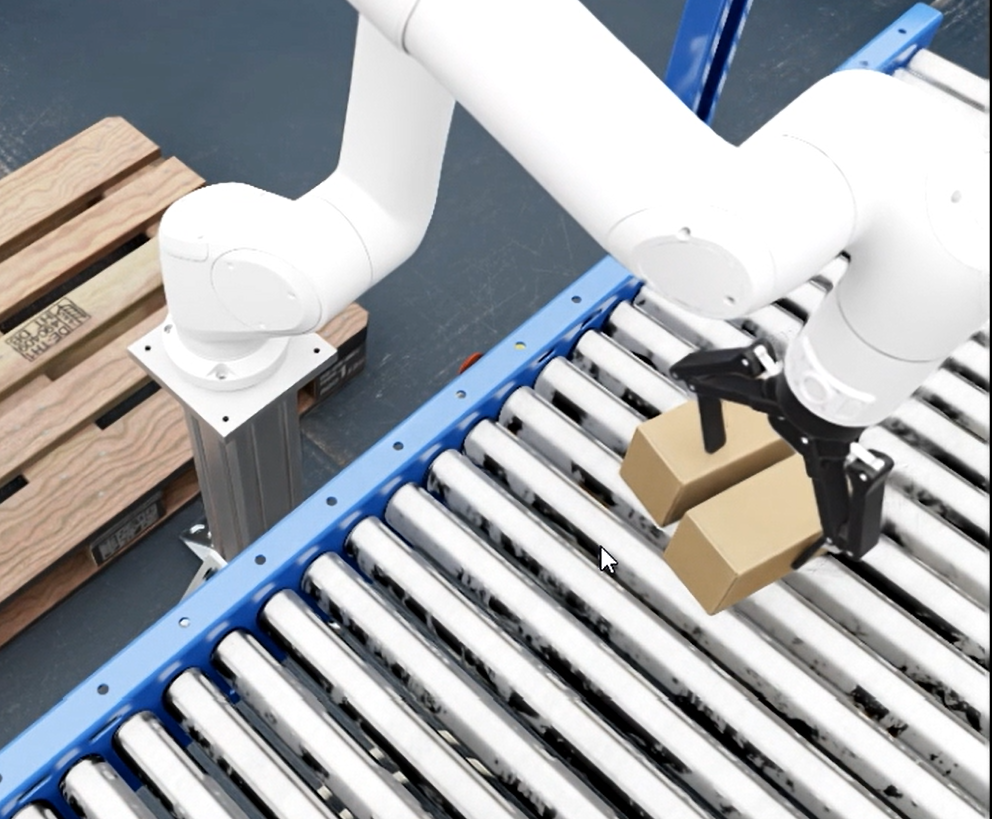}
  \centering
  \caption{Boxes are too close for the gripper.}
  \label{fig:non_repair_examples_uc2}
\end{subfigure}
\caption{Illustration of non-repaired tests}
\label{fig:non_repair_examples}
\end{figure}

For both use cases, most of the failures, more than 99\% were successfully repaired, showing that the fine-tuning of the model improves the vision accuracy, so the soft failures are fixed. As expected, the non-repaired tests result from hard failures, indicating instances where the robot failed to complete the task despite accurate cardbox location predictions by the vision model. Interestingly, we observed a reduction in the number of hard failures after repairing the vision model. This improvement underscores the relationship between enhanced precision in pose estimation and increased system robustness, i.e., a more precise vision model can compensate for certain flaws in the robot control design, thereby contributing to improved task completion rates. This reduction of hard failures by vision model repair is aligned with the feedback of experts on the previous RQ results.% where they confirm the presence of `hidden' soft failures. 
Given the reduced number of non-repaired tests, we conducted a detailed manual analysis to identify two primary root causes for these hard failures. In both use cases, one recurring issue involved the robot arm shaking before placing the box on top of the pallet, resulting in a misaligned placement. An illustration of such a non-repaired test with a misplaced cardboard box is provided in Figure \ref{fig:non_repair_examples_uc2_2}, with a corresponding video demonstration accessible via the provided link \url{https://youtu.be/NzXK4LTLnjo}. In addition, for UC-2, the robot was unable to pick objects when the distance between cardboard boxes was too narrow for the parallel gripper to maneuver around the object, as can be seen in Figure \ref{fig:non_repair_examples_uc2} and its respective scene video at the link \url{https://youtu.be/b23gYYBU9cU}. 
%\url{https://youtu.be/NzXK4LTLnjo}, \url{https://youtu.be/knpesTHjMNM}.

\textbf{Expert feedback.} The generation of failure-revealing test cases is crucial for identifying the design's blind spots and weaknesses. These experiments demonstrate that the suction gripper is generally more suitable than the parallel gripper. As more test cases are simulated, situations arise where the robot cannot perform the pick with the parallel gripper. However, if there is insufficient air pressure for the suction gripper to function properly, several techniques can be implemented to improve the usability of the parallel gripper. For instance, an accelerated conveyor speed can create distance between passing products, or an additional movement can push the product in a certain direction before picking it up if it is in collision with another. Regarding the shaking of the robot, the controller uses Inverse Kinematics (IK) to move the Tool Center Point (TCP) within the Cartesian space. However, this approach is based on numerical optimization, which can be subject to singularity issues, causing the robot to halt its movement. In such situations, the robot controller should move the joints directly without altering the TCP position to escape the singularity and continue the movement. This issue manifests quickly in the simulator as shaking, which affects the placement accuracy.
A possible solution to improve the placement accuracy in this case would be adding a small time delay prior to place allowing the end effector to stabilize. %Revealing test cases with increased risk for singularity in simulation is important for real-world test prioritization. %A potential solution to the revealed singularity problem would be adding a small time delay prior to object placement, allowing the end-effector to stabilize. %Avoiding singularity is a known problem in robotic manipulator control. Revealing test cases with increased risk for singularity in simulation is important for real world test prioritization. %\Dima{Remove this:} In reality, this should not be a problem since real robots have higher precision and can manage such situations more effectively. \Dima{I think we should not say this. On the contrary we should say we will verify if it happens in the real world as part of future work}
\begin{comment}
\begin{tcolorbox}
\textbf{Summary of RQ2:} More than 99 \% of the revealed failures were fixed after fine-tuning the DNN model with the data collected from failures. All the non-repaired tests were found to be due to the controller misconfiguration or environmental interactions. Following the analysis of the non-repaired tests 3 design improvements were proposed by the engineers. 
\end{tcolorbox}
\end{comment}

\subsubsection{RQ3: Sim-to-real ransferability of the in-simulation optimized DNN}
%(Usefulness of the DNN model pre-trained on synthetic data for the real world tasks) How effective is the model pre-trained with synthetic images in making predictions with the corresponding images from the real world?
\hfill \break
\textbf{Motivation.} Although our proposed in-simulation testing effectively identifies ARM design flaws and limitations, our primary objective in optimizing the DNN within the simulation is to facilitate its transfer to the real-world with reduced costs.

\textbf{Method.} We collect and label two limited datasets, consisting of approximately 100 images each, from the actual production environments that conform to the simulation for both use cases. First, we evaluate our original DNN trained with synthetic data, denoted $M_o$, and our optimal, fine-tuned DNN (i.e., obtained by fine-tuning on GA-generated dataset), denoted $M_f$, for UC-1 and UC-2, on the real-world data. We compute the mean Average Precision (mAP), and incorporate the F1-score to quantify the effects of false negatives (i.e., undetected cardboard boxes) and false positives (i.e., irrelevant regions mistakenly identified as cardboard boxes). We evaluate the models on all the real-world data points. Secondly, we divide the real dataset into training (60\%), validation (20\%), and testing (20\%) subsets, then, we used either $M_o$ or $M_f$ as a pre-trained DNN to train the vision model for detecting cardboard boxes in the real-world environment, referring to the resulting models as $M_{or}$ and $M_{of}$, respectively. As a baseline, we train a YOLOv8 model using its standard pre-trained weights exclusively on the real-world training dataset, denoted as $M_r$. We compare the mAP and F1-score of each model on the real world test dataset for each use case, as well as the number of epochs required to converge to the best-fitted model. We consider the model converged if during 5 epochs there is no reduction in loss value. We also report the improvement scores $I_{or}$ and $I_{fr}$, which compare the relative improvement in percentage for mAP and $F_1$ scores for the models $F_o$ and $F_{or}$ as well as $F_f$ and $F_{fr}$. Our intuition is that the performance of the models on the real data should improve as real-world data examples are added to the fine-tuning dataset.

\begin{figure}[h!]
\centering
\begin{subfigure}{0.2\textwidth}
  \includegraphics[scale=0.18]{
  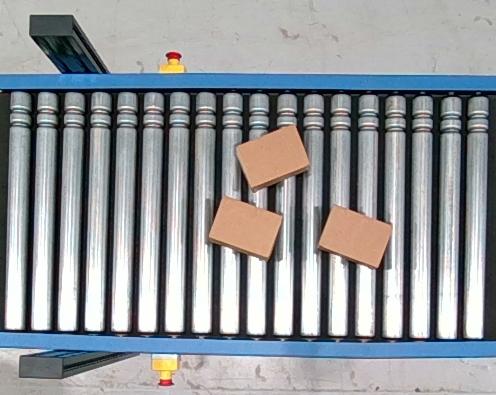}  
  \centering
  \caption{Use case 1 installation}
\end{subfigure}
\begin{subfigure}{0.22\textwidth}
  \includegraphics[scale=0.18]{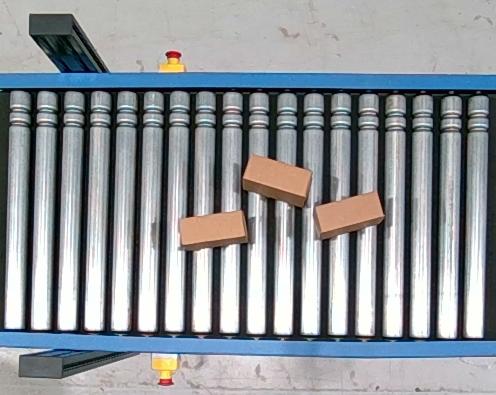}
  \centering
  \caption{Use case 2 installation}
\end{subfigure}
\caption{Real-world setups for both use cases}
\label{fig:real_world}
\end{figure}

\begin{figure}[h!]
\centering
\begin{subfigure}{0.5\textwidth}
  \includegraphics[scale=0.16]{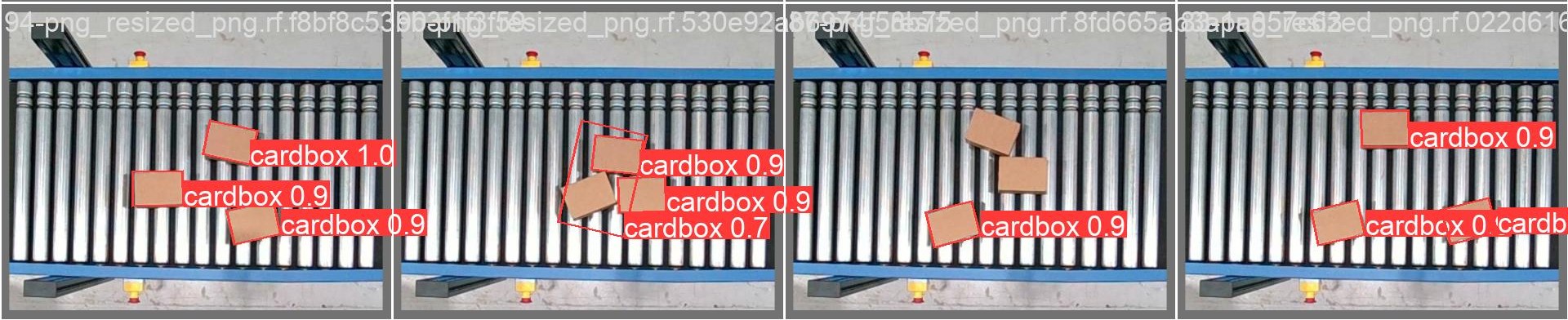}  
  \centering
  \caption{Use case 1 predictions with $M_o$}
\end{subfigure}
\begin{subfigure}{0.5\textwidth}
  \includegraphics[scale=0.16]{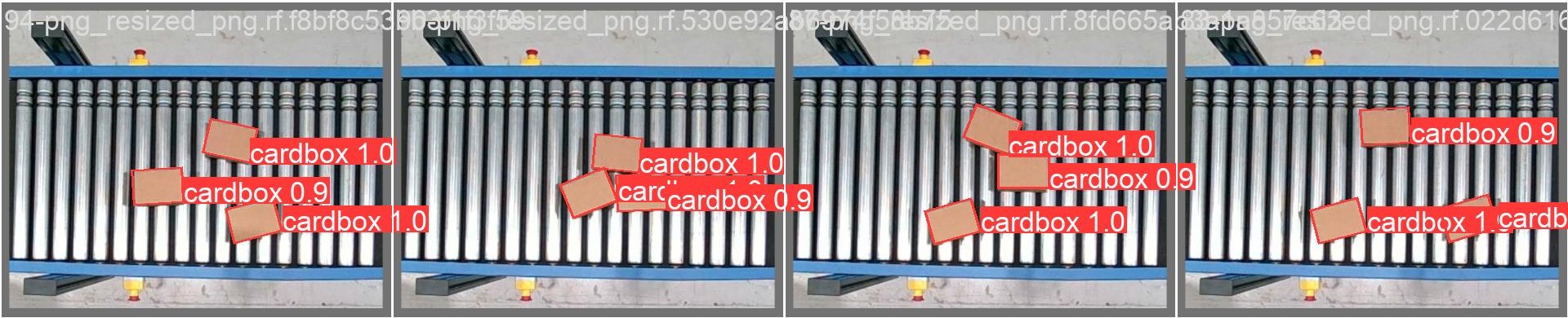}
  \centering
  \caption{Use case 1 predictions with $M_f$}
\end{subfigure}
\begin{subfigure}{0.5\textwidth}
  \includegraphics[scale=0.16]{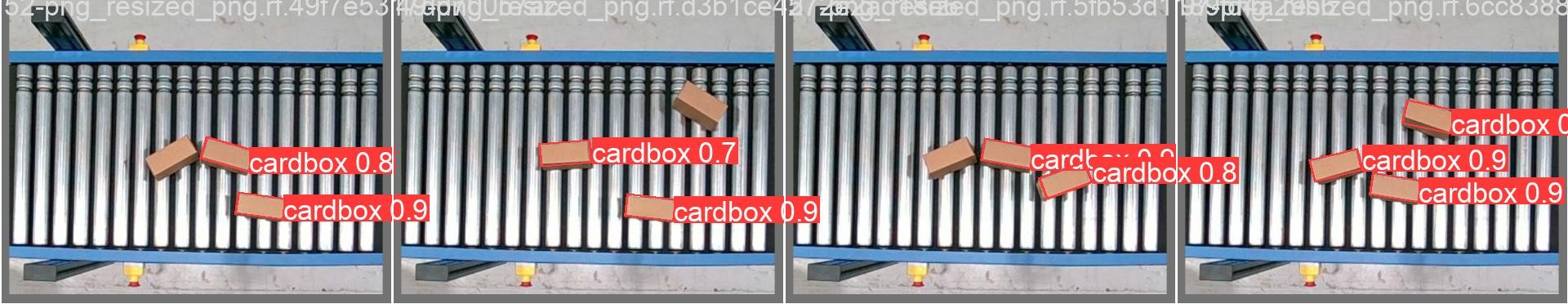}  
  \centering
  \caption{Use case 2 predictions with $M_o$}
\end{subfigure}
\begin{subfigure}{0.5\textwidth}
  \includegraphics[scale=0.16]{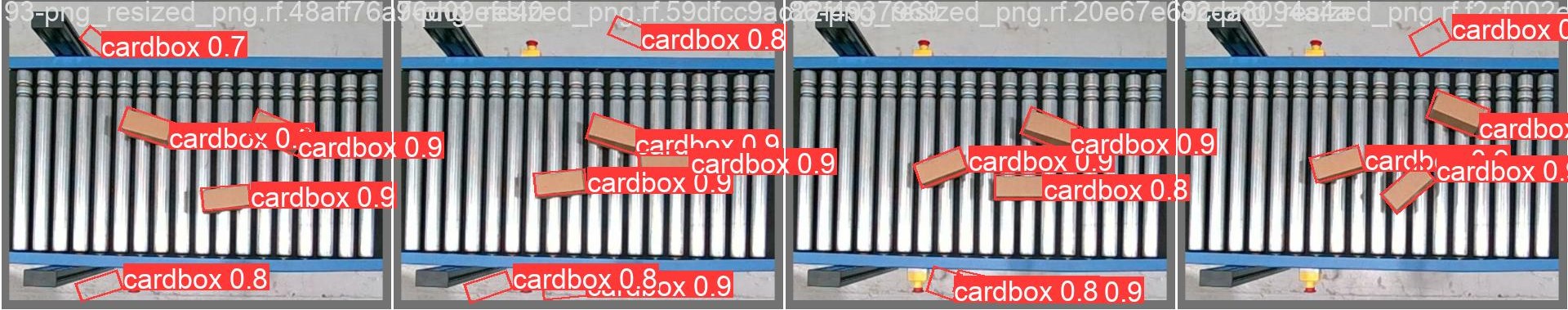}
  \centering
  \caption{Use case 2 predictions with $M_f$}
\end{subfigure}
\caption{Predictions on the real world data with the models trained on synthetic data for use case 1 and use case 2}
\label{fig:real_world_predictions}
\end{figure}

\textbf{Results.} Figure \ref{fig:real_world} shows the images of the scene set up in the real world.
Table \ref{tab:performance1} reports the mAPs and F1-scores obtained by the models, $M_o$ and $M_f$, on the real-world datasets. For UC-1, both models perform well on the real datasets by having performance metrics higher than $0.9$. The fine-tuned model outperforms its original version. As can be seen in Figure \ref{fig:real_world_predictions}, $M_o$ fails to detect some cardboxes, or detects two colliding ones as the same. For UC-2, we found similar results, where $M_f$ outperforms $M_o$, but the overall performance results are substantially lower than the UC-1. As shown in Figure \ref{fig:real_world_predictions}, $M_o$ fails to detect some cardboxes (i.e., false negatives) while $M_f$ misidentifies irrelevant regions as cardboxes (i.e., false positives). Both models struggle to accurately delineate the boundaries of the cardboxes, resulting in a lower mAP compared to UC-1. Despite not being trained on real datasets, these models perform relatively well on real images.%The $M_o$ suffers from underfitting and fails to correctly detect all the cardboxes, even in clearly-visible images. On the other hand, the $M_f$ accurately detects the cardboard boxes but, due to fine-tuning on multiple edge cases, it becomes prone to false detections in some noisy, unseen real observations. 

\begin{table}[h!]
    \centering
    \caption{Performance of the model trained with synthetic data on the real world images}
    \scalebox{0.95}{
    \begin{tabular}{c c c c c}
        \hline
        & \multicolumn{2}{c}{\textbf{Use case 1}} & \multicolumn{2}{c}{\textbf{Use case 2}} \\
        \hline
        & $M_o$ & $M_f$ & $M_o$ & $M_f$ \\
        \hline
        \textbf{mAP} & 0.911 & 0.915& 0.723 & 0.822 \\
        \hline
        \textbf{F1} & 0.962& 0.972 & 0.892 & 0.93 \\
        \hline
        %Precision & 0.989 & 0.983 & 0.947 & 0.887 \\
       % \hline
       % Recall & 0.949 & 0.976 & 0.843 & 0.978 \\
        %\hline
    \end{tabular}}

    \label{tab:performance1}
\end{table}

Table \ref{tab:performance} shows the mAPs, F1-scores, and the number of epochs until convergence for models, $M_r$, $M_{or}$ and $M_{of}$. Despite leveraging an established model like YOLOv8 and its built-in features, such as on-the-fly data augmentation, training an acceptable model with a limited real-world dataset proved challenging. This is indicated in the performance metrics obtained for both use cases in Table \ref{tab:performance}. As introduced in our research work, to achieve better results, a higher investment is necessary to establish an expert-guided protocol for data collection and labelling, ensuring larger real-world datasets with higher coverage for both normal and extreme operating conditions. Notably, for both use cases the fine-tuning on the real world data improved the mAP score compared to training on only synthetic data as seen from the improvement scores $I_{or}$ and $I_{fr}$. \(M_{fr}\) converged on the real-world fine-tuning dataset of UC-1 after only six epochs of  training, whereas \(M_{or}\) failed to achieve similar results even with more epochs. Regarding UC-2, $M_{fr}$ outperforms $M_{or}$ in detection performance and $M_{or}$ shows faster convergence. However, further improvements are necessary to develop a production-ready model for UC-2, as $M_{fr}$ could not surpass a 0.9 mAP.   Discussions with experts yielded the following feedback.

\begin{table}[h!]
    \centering
    \caption{Performance of the model trained on real data without and with pre-trained model on synthetic data}
    \scalebox{0.72}{
    \begin{tabular}{c c c c c c c c c c c c}
        \hline
        & \multicolumn{5}{c}{\textbf{Use case 1}} & \multicolumn{5}{c}{\textbf{Use case 2}} \\
        \hline
        & $M_r$ & \multicolumn{2}{c}{$M_{or}$ } & \multicolumn{2}{c}{$M_{fr}$} & $M_r$ & \multicolumn{2}{c}{$M_{or}$ } & \multicolumn{2}{c}{$M_{fr}$ } \\
        
        &  & $M_{or}$ & $I_{or} (\%)$ & $M_{fr}$ & $I_{fr} (\%)$ &  & $M_{or}$ & $I_{or} (\%)$ & $M_{fr}$ & $I_{fr} (\%)$ \\
        \hline
        \textbf{mAP} & 0.806 & 0.936 & 2.7 & 0.948 & 3.6 & 0.755 & 0.871 & 20.4 & 0.892 & 8.51 \\
        \hline
        \textbf{F1} & 0.91 & 0.968 & 0.62 & 0.998 & 2.67 & 0.905 & 0.924 & 3.58 & 0.958 & 3 \\
        \hline
        %Precision & 0.969 & 0.9947 & 0.972 & 0.925 & 0.954 & 0.944 \\
        %\hline
        % Recall & 0.864 & 0.9324 & 0.972 & 0.886 & 0.895 & 0.971 \\
        %\hline
        \textbf{Epochs} & 26 & 9 & - & 6 & - & 26 & 9 & - & 12 & - \\
        \hline
    \end{tabular}}
    \label{tab:performance}
\end{table}

\begin{comment}
\begin{table}[h!]
    \centering
    \caption{Performance of the model trained on real data without and with pre-trained model on synthetic data}
    \scalebox{0.9}{
    \begin{tabular}{c c c c c c c}
        \hline
        & \multicolumn{3}{c}{Use case 1} & \multicolumn{3}{c}{Use case 2} \\
        \hline
        & $M_r$ & $M_{or}$ & $M_{fr}$ & $M_r$ & $M_{or}$ & $M_{fr}$ \\
        \hline
        MaP & 0.806 & 0.936 & 0.948 & 0.755 & 0.871 & 0.892 \\
        \hline
        F1 & 0.91 & 0.968 & 0.998& 0.905 & 0.924 & 0.958 \\
        \hline
        %Precision & 0.969 & 0.9947 & 0.972 & 0.925 & 0.954 & 0.944 \\
       % \hline
       % Recall & 0.864 & 0.9324 & 0.972 & 0.886 & 0.895 & 0.971 \\
       % \hline
        Epochs & 26 & 9 & 6 & 26 & 9 & 12 \\
        \hline
    \end{tabular}}

    \label{tab:performance}
\end{table}
\end{comment}

\textbf{Expert feedback.} The optimized models in simulation successfully learned relevant inductive biases quickly and efficiently in the production environment. These models were already familiar with the patterns of the cardboxes, the conveyor, and even the colors from the photorealistic simulator (Isaac Sim), as evidenced by their initial performance results on the collected real datasets. Fine-tuning was necessary because of the domain transfer from the simulation to the real world. The models were able to capture the variances of the real domain rapidly, in fewer than 10 epochs, compared to 26 epochs, when no pre-trained model was used. The challenge in solving the real-world version of UC-2 stems from the fisheye effect of the camera. The cardboxes in UC-2 have less width but greater height, creating a fisheye effect that negatively impacts pose estimation by distorting the detection of the box contours. Experts recommend either adjusting the virtual camera parameters in Isaac Sim to simulate a more natural fisheye effect, or modifying the real camera parameters and setup to reduce the fisheye effect on the cardboxes. %This adjustment would minimize discrepancies between the simulation and reality, thereby improving the sim-to-real transferability.
\section{Related work}\label{sec:related_work}
In this section, we discuss the frameworks for testing autonomous robotic systems using simulation. Then, we examine the approaches for improvement of the perception module based on the revealed failures.   
\subsection{Simulation-based Testing}
Several simulation-based testing approaches have been proposed to test end-to-end Autonomous Robotic Systems (ARS) control solutions containing deep learning (DL) components. Examples include DL-driven lane-keeping assist systems tested in the studies by Klikovits et al.~\cite{klikovits2023frenetic}, Zohdinasab et al. \cite{zohdinasab2021deephyperion}, Haq et al.~\cite{haq2020comparing}, and Riccio et al. \cite{riccio2020model}. %Despite these advancements, implementing end-to-end DL-powered solutions in industrial settings remains challenging. In the context of robotic manipulators, robot control is typically implemented using inverse kinematics (IK)-based primitives provided by the robot's software development kit (SDK). Therefore, DL can be employed to process input from sensors and transform it into actionable information that triggers these IK-based primitives to perform context-dependent tasks. For instance, DL vision models can estimate the poses of objects of interest from captured images, enabling the robot to manipulate or avoid them appropriately. Moreover, IK-based control modes are available in simulators supporting ARM applications. 
Despite these advancements, implementing end-to-end DL-powered solutions in industrial settings remains challenging.
Our approach focuses preliminary on testing DL-based perception components within an ARS containing a separate control module, which is based on a DNN or a traditional control algorithm, such as inverse kinematics (IK) based control. %  a simulation environment, integrated with the robot to capture system failures induced by DL and provide early insights into the vision-guided robot system design. 
%Various simulation-based testing approaches have leveraged evolutionary search to evaluate perception and control modules in self-driving automotive applications, such as traffic sign recognition \cite{huai2023doppelganger}, pedestrian detection and emergency braking \cite{ebadi2021efficient}, and general driving tasks \cite{haq2022efficient}. 
 Khatiri et al. \cite{khatiri2023simulation} utilized search-based techniques to systematically generate failure-revealing test cases for autonomous unmanned aerial vehicles (UAVs) with obstacle avoidance vision systems. Following the UAV testing competition \cite{khatiri2024sbft}, several new UAV test generation tools were introduced, including TUMB \cite{tumb}, CAMBA \cite{camba}, WOGAN-UAV \cite{wogan_uav}, DeepHyperion-UAV \cite{deep_hyperion_uav}, and AmbieGen \cite{ambiegen_uav}. 
A relatively small number of research works focuses on testing autonomous collaborative robots. Huck et al. \cite{huck2020simulation} leveraged Monte Carlo Tree Search (MCTS) \cite{lee2020adaptive} to generate high-risk behavior scenarios in the CoppeliaSim simulator \cite{rohmer2013v}. Their goal was to identify human action sequences that could falsify safety requirements, using traditional robot controllers without DL components. Recently, Adigun et al. \cite{adigun2023risk} developed an evolutionary search strategy for testing ARMs with machine learning-based perception components.  %Given the success of evolutionary search in simulation-based testing of autonomous systems, we propose implementing our in-simulation testing for ARM vision components using evolutionary search. This approach will drive test generation towards fault-revealing cases that are relevant for performance assessment and useful for post-testing improvements.
However, this and the aforementioned research works focus on generating the failing test cases, without considering the techniques for improving the system under test from the discovered failures.
%However, the aforementioned research works discusses how revealed failures can be used to improve DL-powered systems. Recently, Lyu et al. \cite{lyu2024search} proposed a technique, named ContrRep, for repairing DNN controllers based on revealed failures. ContrRep assumes that no ground truth data is available for the repair, and it searches for alternative weight values to fix the behavior on failing tests without breaking the passing tests.
%In our approach, as we focus on perception comptestingtestin, the ground truth data is available from the simulation environment. We use a standard DL method for repairing and adapting models: fine-tuning the model on its failures with correct ground truth labels %to adjust its inductive bias and improve its performance in guiding the robot. 
%This is feasible using a simulator that can provide annotated camera images based on the physical and structural environment information.
\subsection{DNN Repair} 
Several techniques have been proposed for repairing DNNs using intelligent data augmentation methods, such as style transfer in DeepRepair \cite{yu2021deeprepair}, search-based data augmentation in SENSEI \cite{gao2020fuzz}, and image-to-image translation of different environmental conditions in TACTIC \cite{li2021testing}. The methods are used in offline mode on 2D images, which can be combined with MARTENS, as we are currently using the on-the-fly augmentation techniques provided by YOLOv8 \cite{ultralytics2024_aug}. It is insufficient to rely solely on 2D image augmentation, and simulation-based test generation is necessary for testing autonomous robotic systems \cite{birchler2024roadmap}. Attaoui et al. \cite{attaoui2024search} developed the DESIGNATE approach, using generative-adversarial networks (GANs) to transform simulator-generated data into realistic images for DNN retraining. Their evaluation on autonomous driving dataset confirmed its effectiveness compared to models trained only on real-world images. In MARTENS, we use a photorealistic simulator to generate images, avoiding the potential biases introduced by GANs. Additionally, we employ online testing to collect retraining images, rather than relying on static scene specifications as in DESIGNATE. 
\section{Conclusions}\label{sec:conclusions}
MARTENS is effective in revealing failures for ARM, allowing the detection of 26\% to 50\% more failures with higher diversity compared to random test generation. Over 99\% of the revealed failures were fixed after fine-tuning the DNN model with data collected from these failures. The remaining failures were attributed to controller misconfiguration or environmental interactions. Following the analysis of these non-repaired tests, engineers proposed three possible design improvements. Using the model repaired with the MARTENS as a pretrained base to perform inference in the real-world environments, we were able to achieve mAPs of 0.948 and 0.892 on test datasets for Use Case 1 and Use Case 2, respectively, after fewer than 10 epochs of fine-tuning. Our future work aims to test the final DNN model on the real robot and develop a method to prioritize and replicate simulation tests for real-world assessment.
%\section{Acknowledgements}
\bibliographystyle{ACM-Reference-Format}
\bibliography{acmart}
\end{document}